\documentclass{article}
\usepackage{arxiv}
\usepackage{amsmath,amsfonts}
\usepackage{algorithmic}
\usepackage{algorithm}
\usepackage{array}
\usepackage[caption=false,font=normalsize,labelfont=sf,textfont=sf]{subfig}
\usepackage{textcomp}
\usepackage{stfloats}
\usepackage{url}
\usepackage{verbatim}
\usepackage{graphicx}
\usepackage{cite}
\usepackage{booktabs}
\usepackage{multirow}
\usepackage[table]{xcolor} % 用于单元格颜色
\usepackage{threeparttable} % 用于表格脚注

\begin{document}

\title{Self-Supervised Path Planning in Unstructured Environments via Global-Guided Differentiable Hard Constraint Projection}

\author{Ziqian Wang
\And 
Chenxi Fang
\And 
Zhen Zhang
\thanks{Corresponding author: Z. Zhang (e-mail: zzhang@tsinghua.edu.cn). Ziqian Wang and Zhen Zhang are with the State Key Laboratory of Tribology in Advanced Equipment, Tsinghua University, Beijing 100084, China and are with the Beijing Key Laboratory of Transformative High-end Manufacturing Equipment and Technology, Department of Mechanical Engineering, Tsinghua University, Beijing 100084, China. Chenxi Fang is with the Automotive Electronics Business Unit at Hirain Inc., China.}
}

\maketitle

\begin{abstract}
Deploying deep learning agents for autonomous navigation in unstructured environments faces critical challenges regarding safety, data scarcity, and limited computational resources. Traditional solvers often suffer from high latency, while emerging learning-based approaches struggle to ensure deterministic feasibility. To bridge the gap from embodied to embedded intelligence, we propose a self-supervised framework incorporating a differentiable hard constraint projection layer for runtime assurance. To mitigate data scarcity, we construct a Global-Guided Artificial Potential Field (G-APF), which provides dense supervision signals without manual labeling. To enforce actuator limitations and geometric constraints efficiently, we employ an adaptive neural projection layer, which iteratively rectifies the coarse network output onto the feasible manifold. Extensive benchmarks on a test set of 20,000 scenarios demonstrate an 88.75\% success rate, substantiating the enhanced operational safety. Closed-loop experiments in CARLA further validate the physical realizability of the planned paths under dynamic constraints. Furthermore, deployment verification on an NVIDIA Jetson Orin NX confirms an inference latency of 94 ms, showing real-time feasibility on resource-constrained embedded hardware. This framework offers a generalized paradigm for embedding physical laws into neural architectures, providing a viable direction for solving constrained optimization in mechatronics. Source code is available at: \url{https://github.com/wzq-13/SSHC.git}.
\end{abstract}

\section{Introduction}
The transition from embodied intelligence to deployable embedded intelligence constitutes a critical frontier in modern mechatronics~\cite{liu2025aligning}. 
In this context, motion planning in unstructured environments stands as a pivotal challenge~\cite{guo2023survey}.
Constrained by irregular obstacles and complex vehicle dynamics~\cite{liu2019minimum}, planners typically struggle to generate trajectories that are both geometrically collision-free and kinematically feasible\cite{chiriatti2021adaptive}.
Effective operation in such dynamic environments often necessitates probabilistic reasoning~\cite{chuang2021novel}. 
However, bridging the gap between these stochastic models and deterministic mechatronic control loops remains a fundamental barrier.
Real-time systems operate under strict temporal deadlines where safe commands must be delivered within the control cycle to maintain closed-loop stability~\cite{zhang2025robust, zheng2024safety}. 
Consequently, the critical gap lies in developing AI agents that provide runtime assurance by generating feasible, smooth, and safe motions with the timing determinism required for embedded deployment.

Traditional motion planning strategies have established a robust foundation for autonomous navigation. Search-based methods, such as Hybrid A*~\cite{HybridAStar}, provide resolution completeness and facilitate global optimality in discretized spaces. Sampling-based algorithms, exemplified by RRT and its variants~\cite{karaman2011sampling, gammell2014informed}, are highly effective at discovering connectivity within high-dimensional geometries through probabilistic completeness. Optimization-based frameworks~\cite{li2022autonomous, guo2025fast, 11030654} have set the standard for generating kinematically smooth and dynamically feasible trajectories. Despite these strengths, deploying these iterative solvers on resource-constrained embedded platforms often requires careful trade-offs between algorithmic convergence rates and real-time response, particularly in unstructured environments with non-convex obstacles.

To address these bottlenecks, data-driven methods like Imitation Learning~\cite{zheng2025diffusion, zhao2024survey} and Deep Reinforcement Learning~\cite{chen2024gaussian, 11247855} offer constant-time inference for embedded intelligence. To enhance the security of learning-based methods, recent research has proposed various approaches, generally falling into two categories: soft regularization and post-hoc shielding. The former incorporates collision penalties or risk-aware terms into the loss function or reward structure~\cite{he2023fear}, encouraging the agent to learn safe behaviors implicitly. However, soft constraints lack deterministic guarantees. The latter employs safety filters, such as Control Barrier Functions (CBFs)~\cite{10923745, 10899401} or predictive safety filters~\cite{10958193,10778608}, to override unsafe actions during execution. Recent advancements have also integrated Koopman operators to enhance the prediction accuracy of such safety governors under complex dynamics\cite{10404071}. While effective at preventing imminent collisions, these distinct add-on modules can lead to conservative behaviors or conflicts with the planner's optimality, potentially inducing deadlocks in cluttered environments.

A distinct stream of research aims to structurally embed rigorous constraints into neural architectures. Broadly, these methods can be categorized into optimization-based approaches that utilize implicit layers or projections~\cite{donti2021dc3, lastrucci2025enforce, iftakher2025physics}, constrained generative sampling within diffusion models~\cite{kurtz2025equality, fishman2023metropolis}, and neuro-symbolic integration that incorporates logical rules~\cite{GIUNCHIGLIA2024109124, 10721277}.
These approaches introduce strong inductive biases, effectively accelerating training convergence and improving the feasibility of generated outputs.
For neuro-symbolic methods, abstracting continuous environmental states into discrete symbols for logical reasoning is difficult in complex, unstructured scenarios.
Conversely, optimization-based projections are well-suited for continuous control, providing runtime assurance by rectifying outputs onto feasible manifolds. However, without high-level guidance, these local solvers struggle to balance strict safety with goal-directed progress. They often prioritize immediate feasibility, risking stagnation in safe but passive local minima~\cite{JIN2019533}.

\begin{figure*}[t]
\centering
\includegraphics[width=\textwidth]{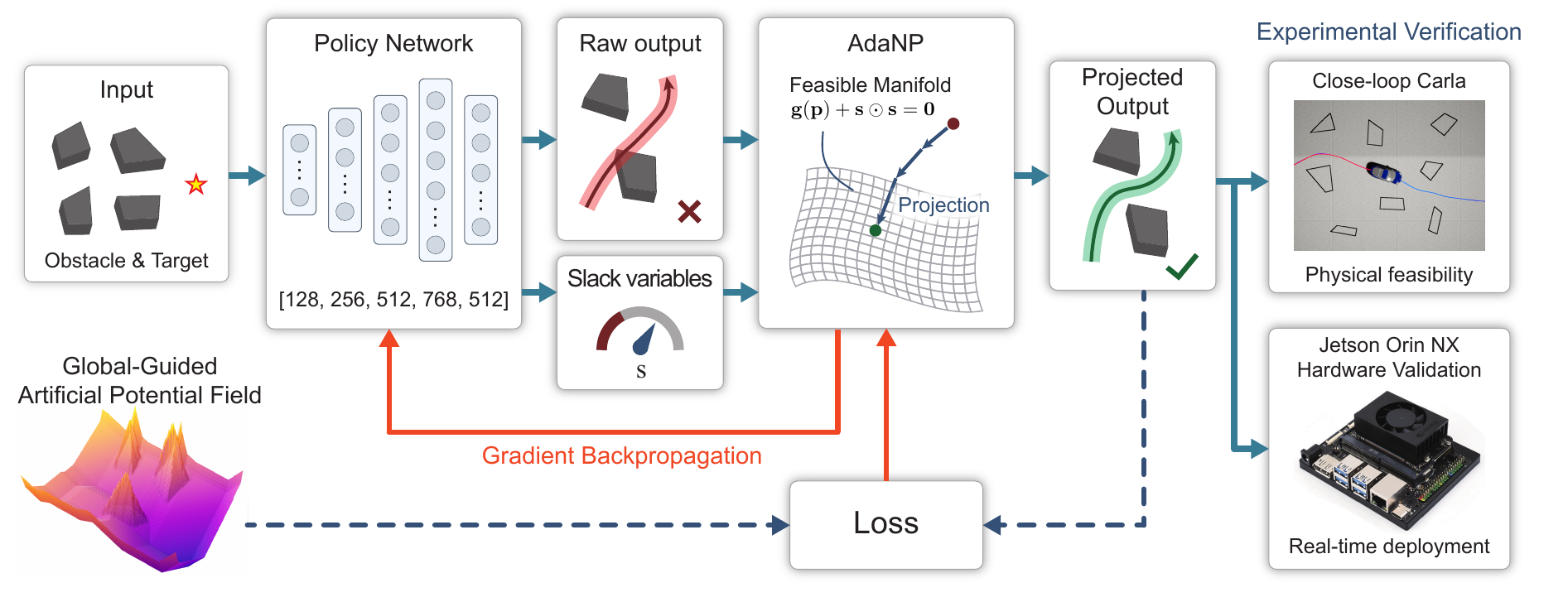}
\caption{Overall framework of the proposed self-supervised planning system. The architecture integrates a Global-Guided Artificial Potential Field to provide dense supervision signals for the Policy Network during the training phase. To ensure safety, an adaptive-depth neural projection (AdaNP) is embedded in the loop, which rectifies the raw network output into a feasible path by solving the constraint equation $\mathbf{g}(\mathbf{p}) + \mathbf{s} \odot \mathbf{s} = \mathbf{0}$.}\label{fig:flow}
\end{figure*}

To bridge this gap, we propose a self-supervised framework that integrates the deterministic constraint handling of differentiable optimization with the global topological guidance of potential fields (shown in Fig. \ref{fig:flow}). We extend the adaptive neural projection mechanism~\cite{lastrucci2025enforce} to handle nonlinear inequality constraints via slack variables, enabling the network to enforce collision avoidance and kinematic boundaries rigorously. Crucially, we couple this with a Global-Guided Artificial Potential Field (G-APF). Serving as a dense supervision signal, the G-APF effectively steers the agent out of local minima, ensuring the projection layer operates on valid initial guesses. This integration ensures that the network not only efficiently generates paths without expert demonstrations but also explicitly enforces the operational safety and feasibility essential for trustworthy embedded mechatronics.

The main contributions of this paper are:
\begin{enumerate}
    \item We propose a self-supervised framework that eliminates reliance on expert demonstrations. A Global-Guided Artificial Potential Field is introduced to provide dense topological supervision, effectively guiding the agent out of local minima and mitigating the data scarcity issue in unstructured navigation.

    \item A differentiable neural projection layer is established to enforce operational safety. It explicitly addresses high-dimensional nonlinear equality and inequality constraints via slack variables and Log-Sum-Exp (LSE) smoothing. It iteratively projects coarse network outputs onto a feasible manifold, acting as a runtime assurance mechanism.

    \item The proposed framework is validated via large-scale benchmarks and closed-loop CARLA~\cite{dosovitskiy2017carla}. The results confirm an 88.75\% success rate and a runtime latency of 94 ms on an NVIDIA Jetson Orin NX. This demonstrates a superior trade-off between navigational robustness and computational efficiency compared to representative baselines, validating the potential for real-time mechatronic applications.
\end{enumerate}
\section{Problem Formulation}\label{sec:problem_formulation}

We define the path planning task as generating a sequence of discrete waypoints $\mathcal{P} = \{ \mathbf{p}_0, \mathbf{p}_1, \dots, \mathbf{p}_T \}$ in a 2D plane, where $\mathbf{p}_t = {[x_t, y_t]}^T$. The planner must generate a path connecting the robot's current state and the target area, while optimizing efficiency, smoothness, and avoiding collisions.

\subsection{Constraints Formulation}

\subsubsection{Waypoint Density}
To guarantee path resolution, the Euclidean distance between consecutive waypoints is constrained by a maximum threshold $d_{\text{max}}$:
\begin{equation}
    \| \mathbf{p}_{t+1} - \mathbf{p}_t \|_2 \leq d_{\text{max}}, \quad \forall t \in \{0, \dots, T-1\}.
\end{equation}

\subsubsection{Kinematic Constraints}
The vehicle's steering capability is limited by its physical design. The maximum allowable curvature $\kappa_{\text{max}}$ is determined by the maximum steering angle $\delta_{\text{max}}$ and the wheelbase $L$:
\begin{equation}
    \kappa_{\text{max}} = \tan(\delta_{\text{max}}) / L.
\end{equation}

For the discrete path $\mathcal{P}$, we calculate the local curvature $\kappa_t$ at waypoint $\mathbf{p}_t$ using the Menger curvature, which is the inverse of the radius of the circumcircle passing through $\mathbf{p}_{t-1}, \mathbf{p}_t$, and $\mathbf{p}_{t+1}$.  The constraint requires:
\begin{equation}
    |\kappa_t(\mathbf{p}_{t-1}, \mathbf{p}_t, \mathbf{p}_{t+1})| \leq \kappa_{\text{max}}, \quad \forall t \in \{1, \dots, T-1\}.
    \label{eq:curvature_constraint}
\end{equation}

\subsubsection{Collision Avoidance}
The vehicle must strictly avoid the set of static obstacles $\mathcal{O}$. For every waypoint $\mathbf{p}_t$, the distance to the nearest obstacle boundary must exceed the safety radius $r_\text{safe}$:
\begin{equation}
    \min_{\mathbf{o} \in \mathcal{O}} \| \mathbf{p}_t - \mathbf{o} \|_2 \geq r_\text{safe}, \quad \forall t.
\end{equation}

\subsection{Optimization Objective}
The planning framework aims to minimize the total travel distance $J_{\text{len}}$ and the final deviation from the target $J_{\text{goal}}$, defined as:
\begin{align}
    J_{\text{len}} &= \sum_{t=0}^{T-1} \| \mathbf{p}_{t+1} - \mathbf{p}_t \|_2, \\
    J_{\text{goal}} &= \| \mathbf{p}_T - \mathbf{p}_{\text{target}} \|_2^2.
\end{align}

\section{Methodology}\label{sec:method}

In this section, we present the proposed self-supervised planning framework. As illustrated in Fig.~\ref{fig:flow}, the system comprises three core components: a Global-Guided Potential Field that generates dense supervision signals, a Differentiable Obstacle Representation module that smooths collision constraints, and a Hard Constraint Projection Layer that enforces kinematic and safety feasibility.

\subsection{Global-Guided Potential Field}
\begin{figure}[htbp]
    \centering
    \includegraphics[width=0.40\textwidth]{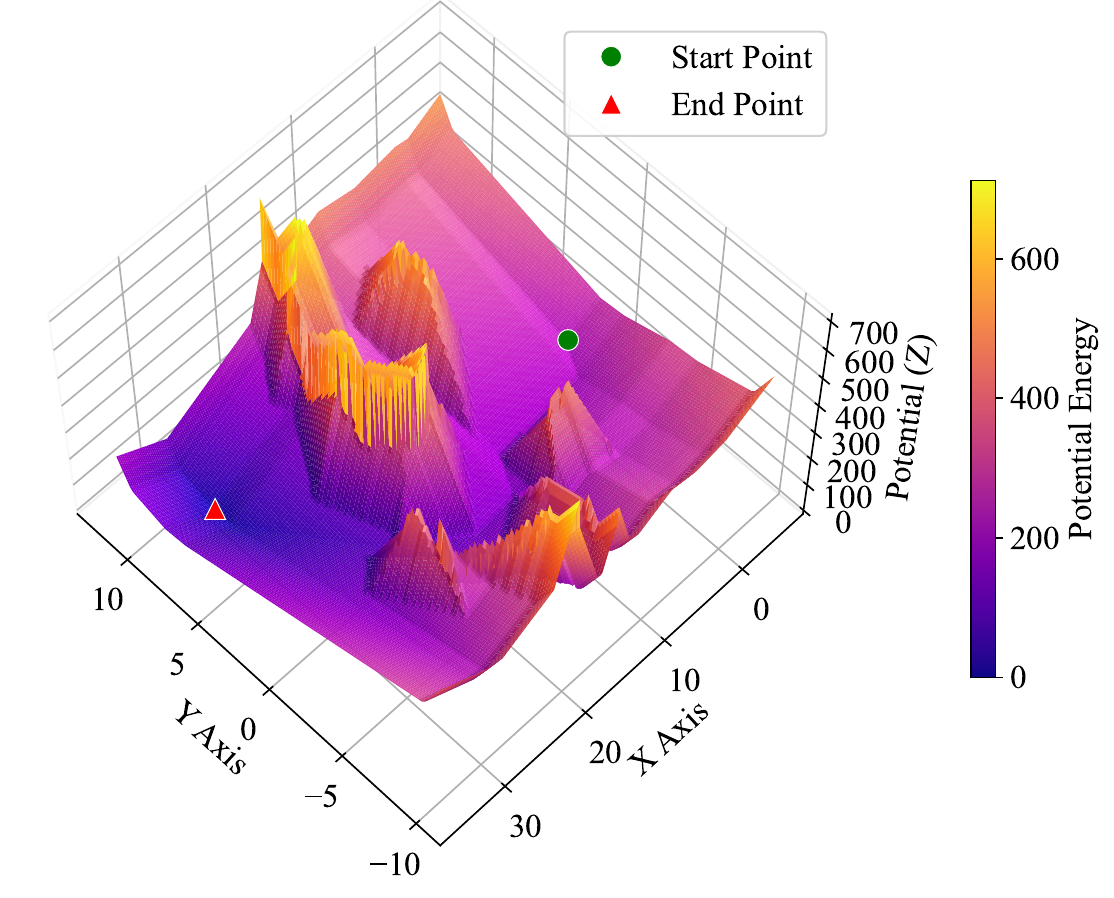}
    \caption{Visualization of the Global-Guided Potential Field. The multi-source wavefront propagation creates a convex valley along the optimal path, providing dense gradient guidance to steer the agent out of local minima.}
    \label{fig:potential_field}
\end{figure}

To guide the planned path toward the target while maintaining topological consistency, we construct a dense guidance field $V(\mathbf{u})$ over a grid with 0.2~m resolution (shown in Figure~\ref{fig:potential_field}). This field serves as a pre-computed supervision signal that provides continuous supervisory signals for the policy network throughout the training process.

First, a global reference path $\mathcal{P}^*$ is generated via Dijkstra's algorithm\cite{dijkstra1959note}. We initialize the potential values on $\mathcal{P}^*$ based on their geodesic distance to the goal.
Subsequently, we employ a Multi-Source Wavefront Propagation to expand the potential field from $\mathcal{P}^*$ into the free space. The potential value of a node $\mathbf{u}_{\text{next}}$ is iteratively updated from its neighbor $\mathbf{u}_{\text{curr}}$:
\begin{equation}
    V(\mathbf{u}_{\text{next}}) = \min \left( V(\mathbf{u}_{\text{next}}), \: V(\mathbf{u}_{\text{curr}}) + \|\Delta \mathbf{u}\|_2 + V_{\text{obs}}(\mathbf{u}_{\text{next}}) \right),
    \label{eq:wavefront}
\end{equation}
where $\|\Delta \mathbf{u}\|_2$ denotes the Euclidean distance. The term $V_{\text{obs}}(\mathbf{u})$ imposes a high penalty constant within obstacle regions, creating a steep repulsive gradient that drives the optimization toward the safe valley centered on $\mathcal{P}^*$. This mechanism drastically elevates the potential values within obstacle areas, creating a steep repulsive gradient at the boundaries that effectively pushes the optimization process away from collisions and toward the safe free space.

\begin{figure}[htbp]
\centering
\subfloat[]{\includegraphics[width=0.20\textwidth]{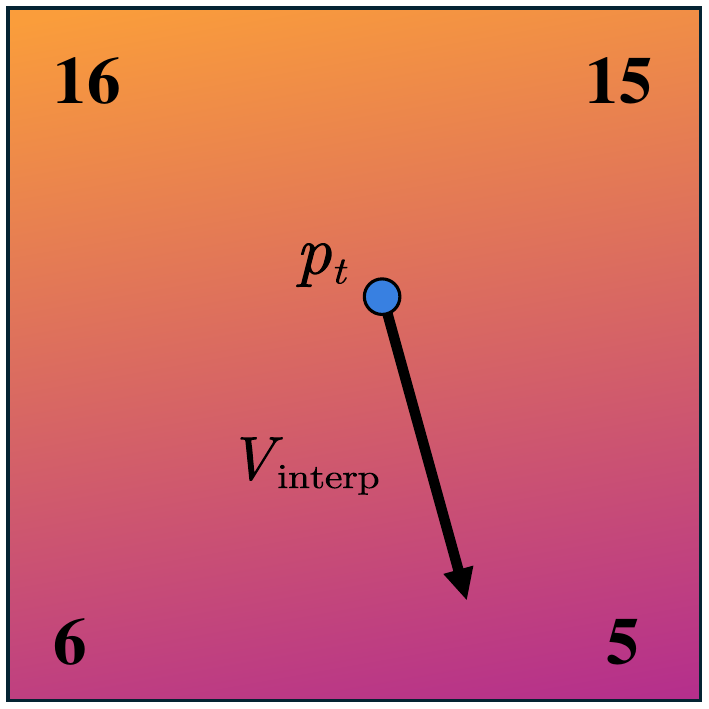}\label{fig_first_case}}
\quad
\subfloat[]{\includegraphics[width=0.20\textwidth]{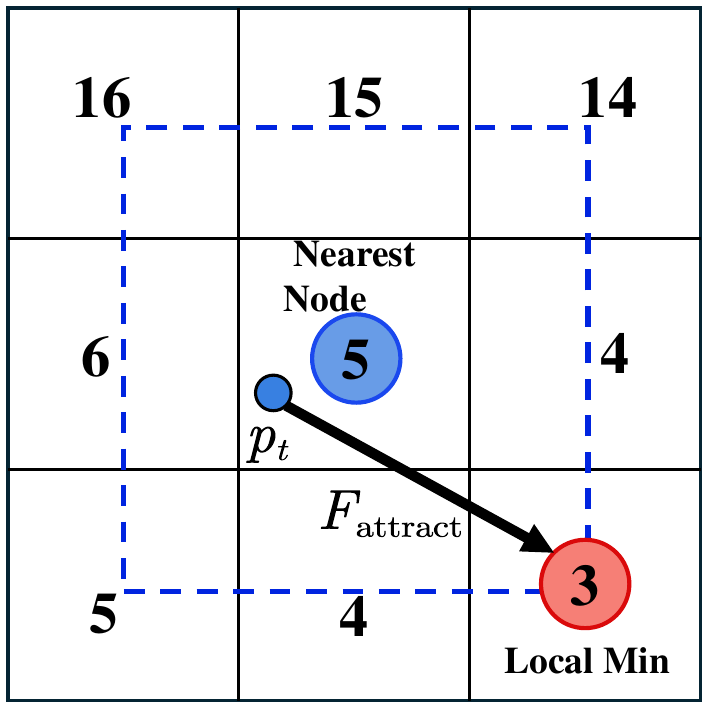}\label{fig_second_case}}
\caption{Global-Guided Potential Field Loss.\ (a) Direct potential minimization guides the agent toward lower potential regions.\ (b) Local attraction ensures continuous progress along the potential valley, especially in obstacle-dense areas.}\label{fig:potential_loss}
\end{figure}

To drive the agent toward the goal while avoiding local minima, the potential field loss $\mathcal{L}_{\text{pot}}$ combines a global gradient term with a local attraction term:
\begin{equation}
    \mathcal{L}_{\text{pot}} = \frac{1}{T} \sum_{t=1}^{T} \left( V_{\text{interp}}(\mathbf{p}_t) + \beta \min_{\mathbf{u} \in \mathcal{N}(\mathbf{n}_t)} \| \mathbf{p}_t - \mathbf{u} \|_2^2 \right).
\end{equation}
The first term aligns the path with the global energy valley via bilinear interpolation. The second term ($\beta > 0$) pulls the agent toward the local minimum node $\mathbf{u}$ within $\mathcal{N}(\mathbf{n}_t)$, preventing stagnation or oscillations in obstacle-dense regions where the first term's gradient may become nearly orthogonal to the desired direction of progress (shown in Figure~\ref{fig:potential_loss}).

\subsection{Differentiable Hard Constraint Projection}
We embed hard physical constraints into the network architecture by parameterizing slack variables within the output layer, enabling an end-to-end differentiable projection for constrained optimization.

\subsubsection{Differentiable Collision Representation}
To facilitate gradient-based optimization within the projection layer, we approximate the non-differentiable collision boundaries using the LSE technique~\cite{boyd2004convex}. Unlike discrete Boolean checks, LSE transforms the collision condition into a smooth, differentiable manifold.

For the $j$-th obstacle defined by the intersection of $M_j$ half-planes $l_{j,m}(\mathbf{p}) \ge 0$, the signed distance is approximated by $c_j(\mathbf{p}) \approx -\frac{1}{\alpha}\log(\sum_{m=1}^{M_j} e^{-\alpha l_{j,m}(\mathbf{p})})$. The global collision constraint aggregates all $N_{\text{obs}}$ obstacles via the smooth maximum operator:
\begin{equation}
    C_{\text{obs}}(\mathbf{p}) = \text{LSE}_{\text{max}}(\{c_j(\mathbf{p})\}_j) =\frac{1}{\alpha}\log\left(\sum_{j=1}^{N_{\text{obs}}} e^{\alpha c_j(\mathbf{p})}\right),
    \label{eq:lse_collision}
\end{equation}
The feasibility requires $C_{\text{obs}}(\mathbf{p}) < 0$. Here, $\alpha$ is a positive scaling factor controlling the approximation accuracy. Theoretically, as $\alpha \to \infty$, the LSE function converges to the exact max/min operator. However, excessively large values of $\alpha$ induce sharp transitions in the constraint landscape, resulting in exploding gradients that can destabilize the optimization process. We empirically selected $\alpha=10$ to strike a balance between geometric fidelity and numerical stability. Visualization is in Supplementary Material.

\subsubsection{Differentiable Constraint Transformation}
To enable gradient-based projection, we reformulate the inequality constraints defined in Section \ref{sec:problem_formulation} into equality constraints $\mathbf{h}(\mathbf{y}) = \mathbf{0}$ by introducing augmented slack variables $\mathbf{s}$. The state vector at time $t$ is expanded to $\mathbf{y}_t = [\mathbf{p}_t^T, \mathbf{s}_t^T]^T$. By enforcing $\mathbf{g}(\mathbf{p}) + \mathbf{s} \odot \mathbf{s} = \mathbf{0}$, the inequality $\mathbf{g}(\mathbf{p}) \leq 0$ is strictly satisfied while maintaining differentiability.

We aggregate the constraints into a unified vector $\mathbf{h}(\mathbf{y}_t)$. For collision avoidance, the vehicle footprint is approximated by three covering circles with centers $\mathbf{c}_{t,k}, (k \in \{1, 2, 3\})$ to ensure differentiability. Kinematic and density constraints use squared terms for numerical smoothness:
\begin{equation}
    \mathbf{h}(\mathbf{y}_t) = 
    \begin{bmatrix}
        \text{LSE}_{\text{max}}\left(\{C_{\text{obs}}(\mathbf{c}_{t,k}) + r_{\text{safe}}\}_{k=1}^{3}\right) + s_{t,\text{obs}}^2 \\
        \kappa_t^2 - \kappa_{\text{max}}^2 + s_{t,\kappa}^2 \\
        \| \mathbf{p}_{t+1} - \mathbf{p}_t \|_2^2 - d_{\text{max}}^2 + s_{t,d}^2
    \end{bmatrix} 
    = \mathbf{0}.
    \label{eq:unified_constraints}
\end{equation}
By concatenating these terms over the horizon $T$, we define the unified constraint problem $\mathbf{h}(\mathbf{y}) = \mathbf{0}$, which serves as the optimization target for the subsequent projection layer.

\subsubsection{Differentiable Projection Layer}
We employ a differentiable projection layer based on the Newton-Raphson method to project the coarse network output $\mathbf{y}_k$ onto the feasible manifold defined by $\mathbf{h}(\mathbf{y}) = \mathbf{0}$. This is formulated as a Quadratic Programming (QP) problem:
\begin{equation}
    \begin{aligned}
    \mathbf{y}_{k+1} = \quad & \underset{\mathbf{y}}{\text{argmin}} \quad \frac{1}{2} \| \mathbf{y} - \mathbf{y}_k \|_2^2 \\
    & \text{s.t.} \quad \mathbf{h}(\mathbf{y}_k) + \mathbf{J}_{\mathbf{h}}(\mathbf{y}_k) (\mathbf{y} - \mathbf{y}_k) = \mathbf{0},
    \end{aligned}
\end{equation}
where $\mathbf{J}_{\mathbf{h}}$ is the Jacobian matrix. Solving the Lagrangian dual yields the closed-form minimum-norm Newton step:
\begin{equation}
    \mathbf{y}_{k+1} = \mathbf{y}_k - \mathbf{J}_{\mathbf{h}}^\dagger (\mathbf{y}_k) \mathbf{h}(\mathbf{y}_k),
\end{equation}
with the pseudo-inverse $\mathbf{J}_{\mathbf{h}}^\dagger = \mathbf{J}_{\mathbf{h}}^T (\mathbf{J}_{\mathbf{h}} \mathbf{J}_{\mathbf{h}}^T)^{-1}$. This update is applied iteratively until $\|\mathbf{h}(\mathbf{y}_k)\|_\infty < \epsilon$, or the number of steps reaches the maximum iteration limit ($I_\text{max}$). Enforcing a strict iteration cap guarantees deterministic worst-case execution time. Since all operations are differentiable, gradients from $\mathbf{y}^*$ flow to the policy, enabling end-to-end training. To ensure real-time feasibility on embedded GPUs, we optimize the projection solver via static computation graphs and kernel fusion.

\subsection{Curriculum Training Strategy}
Solving a highly non-convex optimization problem with strict hard constraints from scratch typically leads to severe solver oscillation and gradient instability. To mitigate this, we design a two-stage curriculum learning strategy\cite{bengio2009curriculum} that progressively increases the difficulty of the constraints.

\subsubsection{Stage 1: Soft-Constraint Training}
In this stage, the hard projection layer is deactivated. The network learns the global topology and rough path feasibility using soft loss.

The constraints defined in the previous section are relaxed into penalty terms using the ReLU function:
\begin{align}
    \mathcal{L}_{\text{soft\_obs}} &= \frac{1}{T} \sum_{t,k} \text{ReLU}(C_{\text{obs}}(\mathbf{c}_{t,k}) + r_\text{safe}), \\
    \mathcal{L}_{\text{soft\_curve}} &= \frac{1}{T} \sum_{t} \text{ReLU}(|\kappa_t| - \kappa_{\text{max}}), \\
    \mathcal{L}_{\text{soft\_dist}} &= \frac{1}{T} \sum_{t} \text{ReLU}(\| \mathbf{p}_{t+1} - \mathbf{p}_t \|_2 - d_{\text{max}}).
\end{align}

To facilitate the transition to the hard-constraint projection in Stage 2, the network must learn to predict appropriate slack variables $\mathbf{s}$ that satisfy the constraints. We introduce an auxiliary loss $\mathcal{L}_{\text{slack}}$:
\begin{equation}
\mathcal{L}_{\text{slack}} = \frac{1}{T} \sum_{t=1}^{T} \left| \mathbf{h}(\text{sg}(\mathbf{p}_t), s_{t,\text{obs}}, s_{t,\kappa}, s_{t,d}) \right|,
\end{equation}
where $\text{sg}(\cdot)$ denotes the stop-gradient operator. Minimizing $\mathcal{L}_{\text{slack}}$ effectively calibrates the slack variables to the environment's topology. This mitigates excessive corrections during projection, thereby reducing the risk of solver oscillation or convergence to suboptimal local minima.

Let $\mathcal{L}_{\text{soft}}$ denote the aggregated soft constraint penalty:
\begin{equation}
    \mathcal{L}_{\text{soft}} = \mathcal{L}_{\text{soft\_obs}} + \mathcal{L}_{\text{soft\_curve}} + \mathcal{L}_{\text{soft\_dist}}.
\end{equation}
The total loss for Stage 1 is then formulated as:
\begin{equation}
    \mathcal{L}_{\text{Stage1}} = \mathcal{L}_{\text{pot}} + \mathcal{L}_{\text{slack}} + \lambda_{\text{soft}} \mathcal{L}_{\text{soft}}.
\end{equation}

\subsubsection{Stage 2: Hard-Constraint Projection}
Once the network converges to a preliminary solution, we activate the differentiable projection layer. This layer projects the coarse prediction onto the feasible manifold defined by $\mathbf{h}(\mathbf{y})=\mathbf{0}$. 

\begin{figure*}[t]
    \centering
    
    \begin{minipage}{\textwidth}
        \centering
        \includegraphics[width=0.9\linewidth]{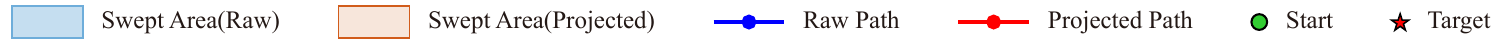}
    \end{minipage}

    \begin{tabular}{c c c}
        \subfloat[Stage 1: Epoch 100]{\includegraphics[width=0.3\textwidth]{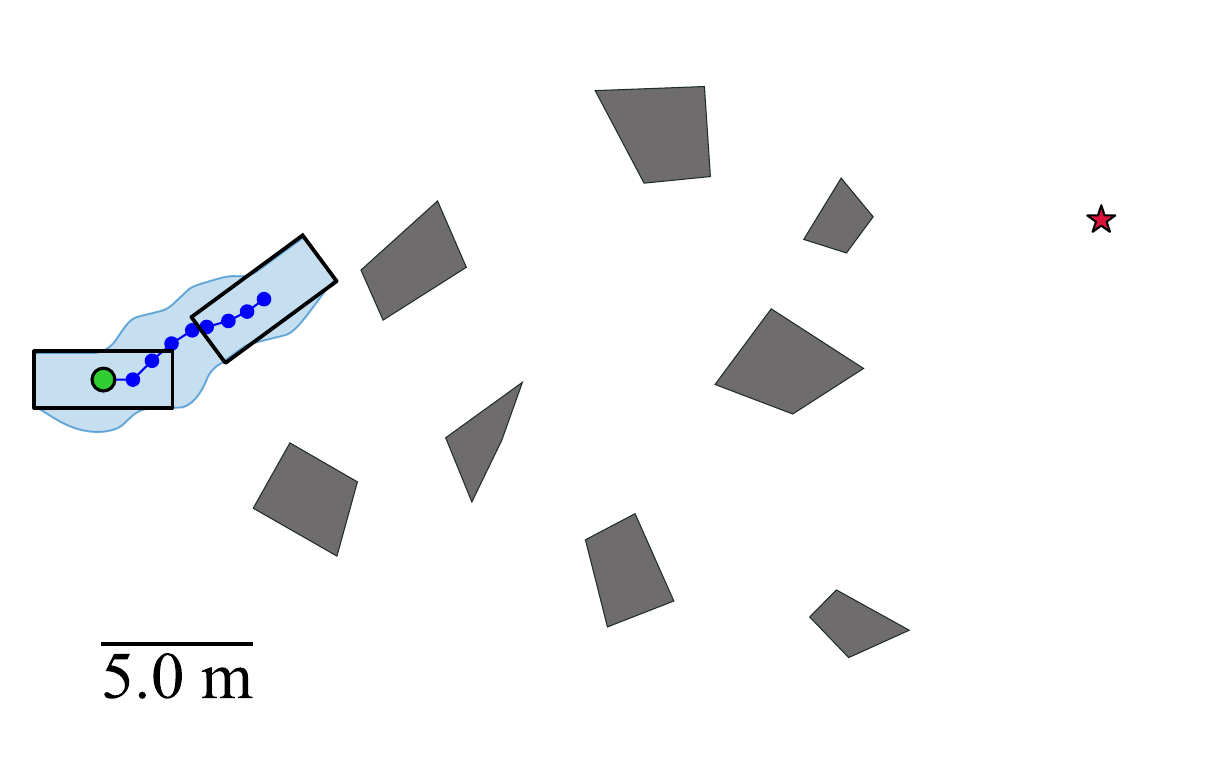}} &
        \subfloat[Stage 1: Epoch 200]{\includegraphics[width=0.3\textwidth]{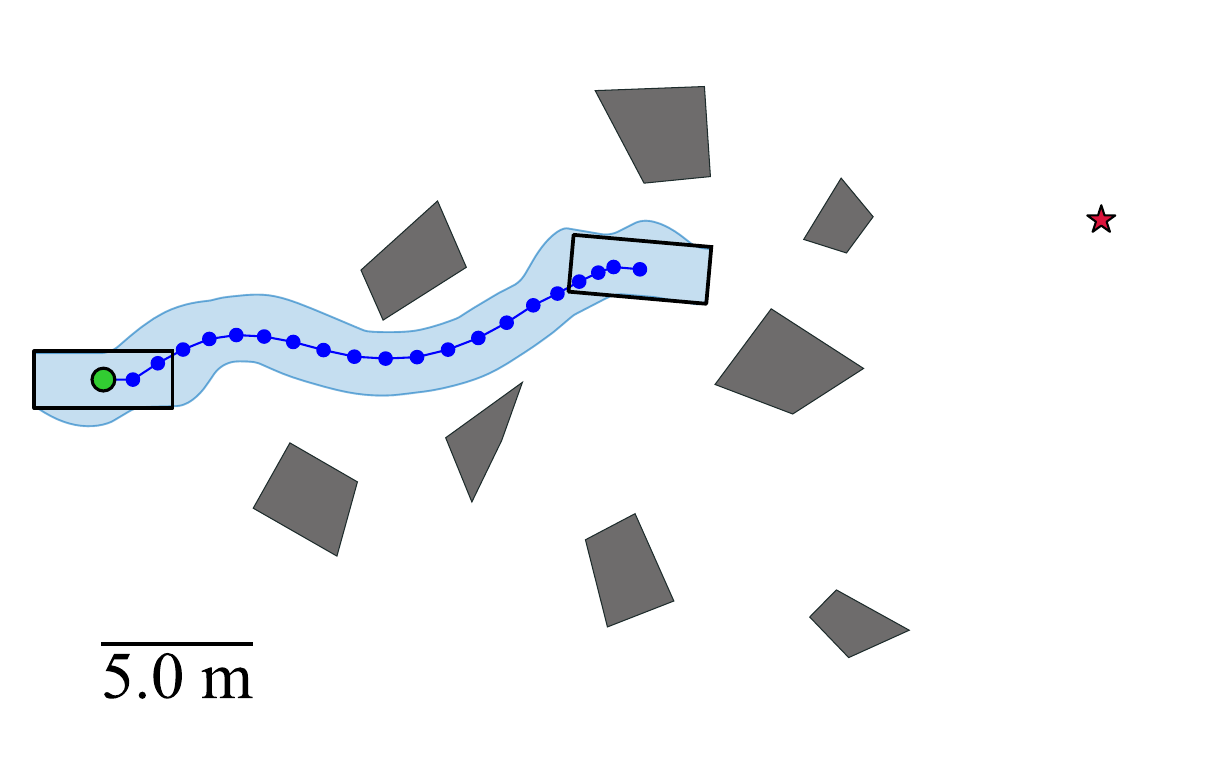}} &
        \subfloat[Stage 1: Epoch 400]{\includegraphics[width=0.3\textwidth]{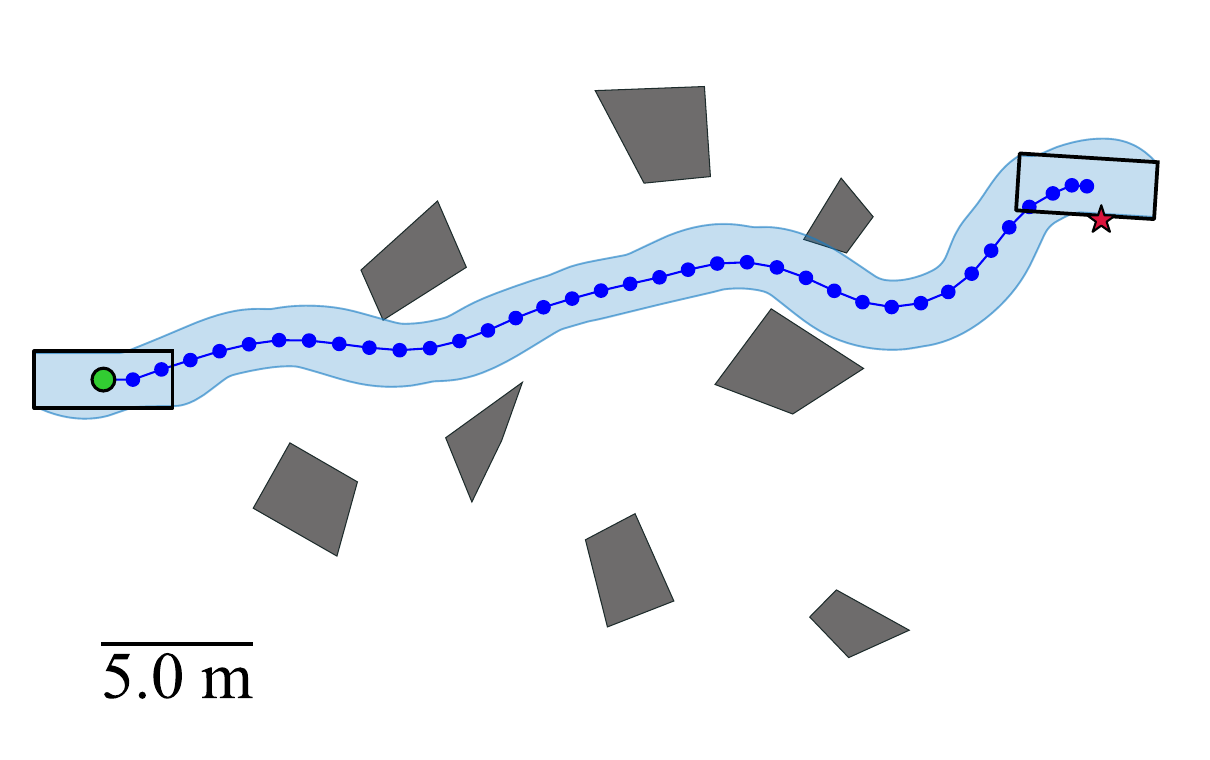}} \\
        \subfloat[Stage 2: Epoch 0]{\includegraphics[width=0.3\textwidth]{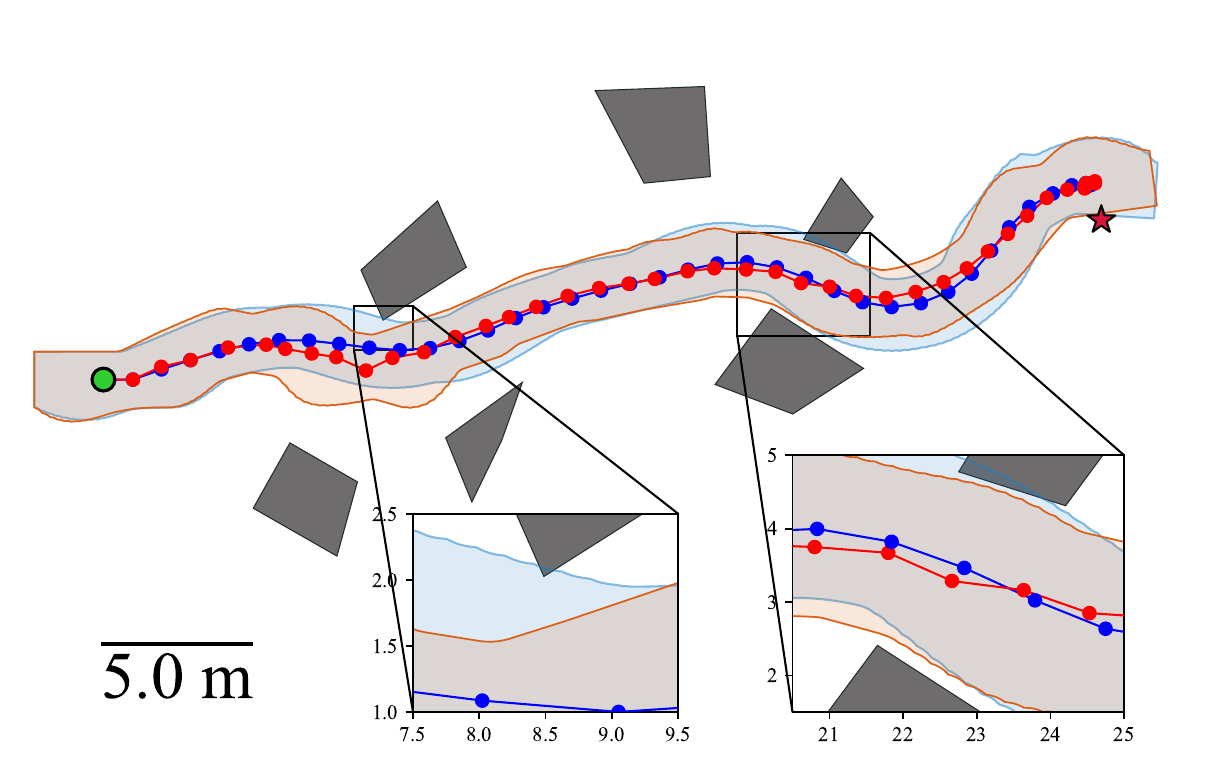}} &
        \subfloat[Stage 2: Epoch 20]{\includegraphics[width=0.3\textwidth]{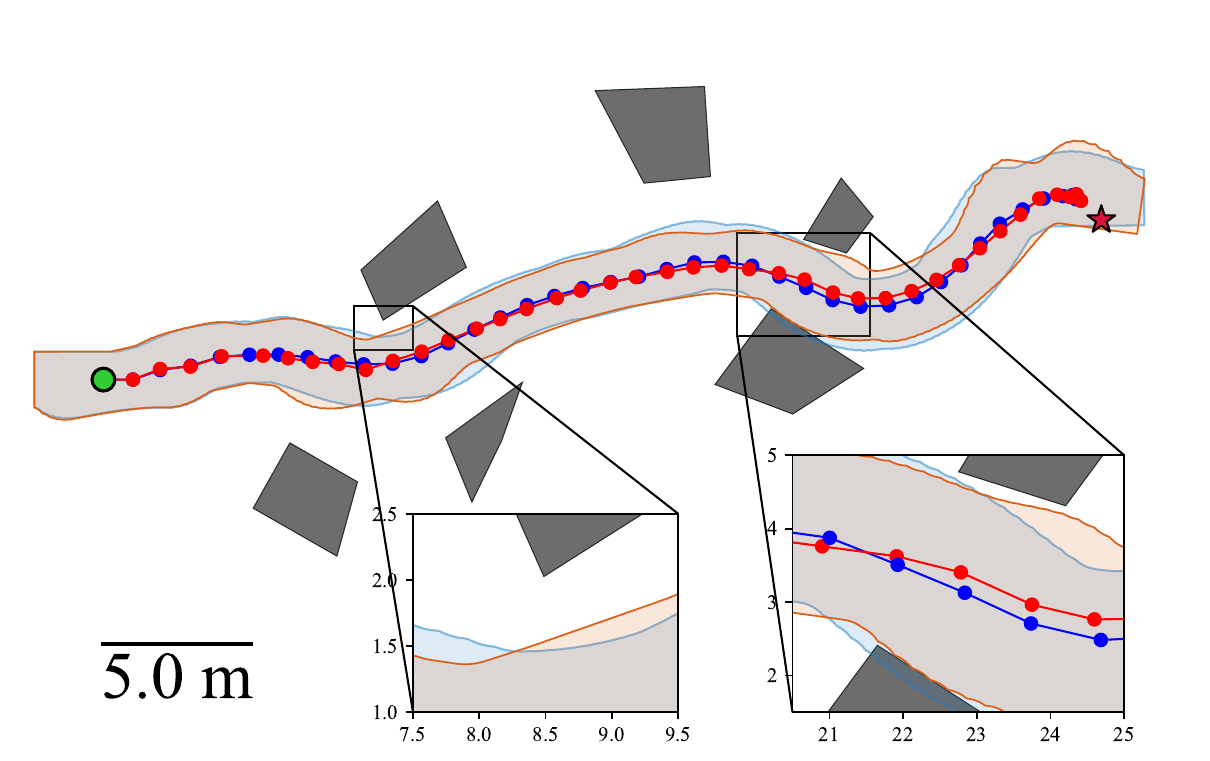}} &
        \subfloat[Stage 2: Epoch 40]{\includegraphics[width=0.3\textwidth]{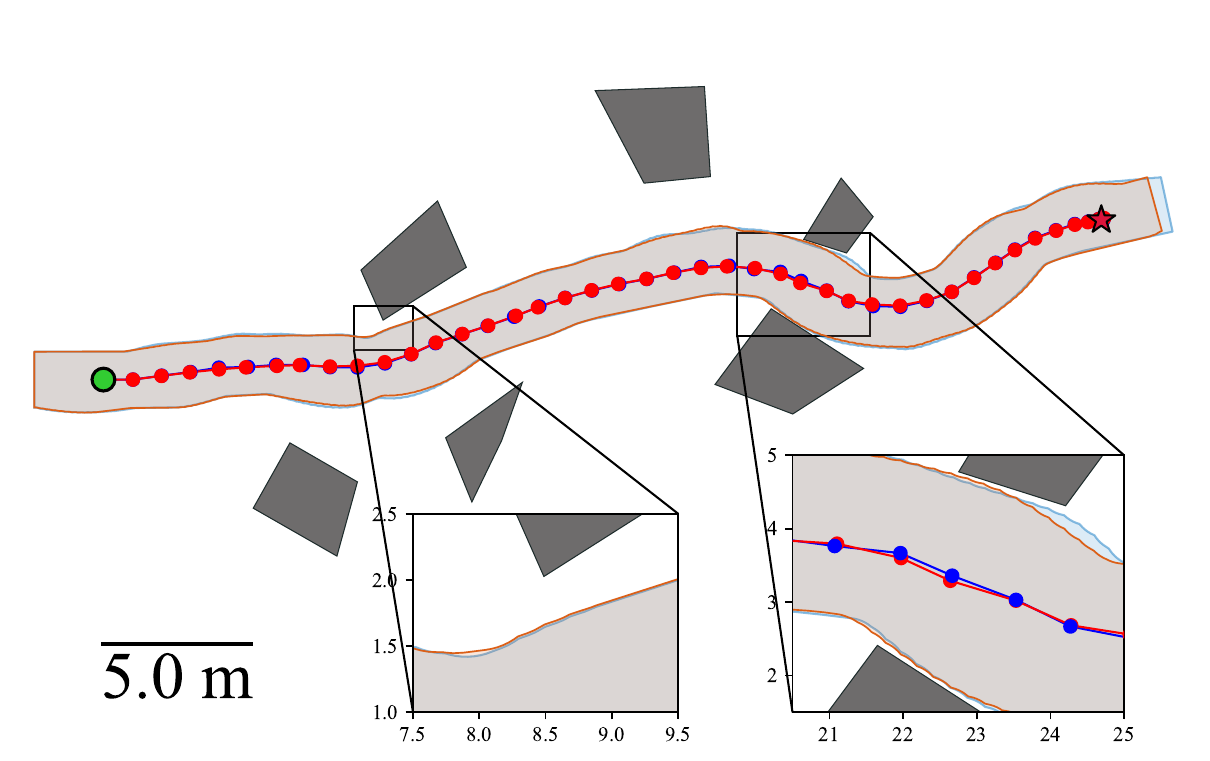}}
    \end{tabular}

    \caption{The evolution of the planned path during training. (a)-(c) Stage 1: The network learns to follow the potential valley with soft constraints roughly. (d)-(f) Stage 2: The hard projection layer refines the path to enhance safety and kinematic feasibility.}
    \label{fig:process}
\end{figure*}

In this stage, the explicit soft constraint losses and the slack estimation loss are removed. We introduce a projection loss $\mathcal{L}_{\text{proj}}$ to minimize the correction magnitude. The objective becomes:
\begin{equation}
    \mathcal{L}_{\text{Stage2}} = \mathcal{L}_{\text{pot}} + \lambda_{\text{proj}} \underbrace{\frac{1}{T} \sum_{t=0}^{T} \| \hat{\mathbf{y}}_t - \mathbf{y}_t \|_2^2}_{\mathcal{L}_{\text{proj}}}.
\end{equation}
where $\hat{\mathbf{y}}_t$ is the raw network output and $\mathbf{y}_t$ is the projected feasible output. Minimizing $\mathcal{L}_{\text{proj}}$ aligns the raw output $\hat{\mathbf{y}}_t$ with the feasible projection $\mathbf{y}_t$. This regularization anchors predictions within the solver's convergence basin, effectively treating the projection as a lightweight local refinement and significantly reducing inference latency.
The evolution of the path through these stages is visualized in Fig.~\ref{fig:process}. The complete training procedure is outlined in Algorithm \ref{alg:training_process}.

\begin{algorithm}[htbp]
\caption{Two-Stage Self-Supervised Training Strategy}
\label{alg:training_process}
\begin{algorithmic}
\REQUIRE Dataset $\mathcal{D}$, Pre-computed Potential Field $V(\mathbf{u})$, Weights $\lambda_{\text{soft}}, \lambda_{\text{proj}}$, Epochs $E_{S1}, E_{S2}$.
\ENSURE Safety-Critical Policy Network $\pi_\theta$.
\STATE \textbf{Initialize} $\pi_\theta$ randomly.
\FOR{epoch $e = 1$ to $E_{S1}$}
    \FOR{batch $B \in \mathcal{D}$}
        \STATE $\hat{\mathbf{y}} = [\mathbf{x}, \mathbf{s}] \leftarrow \pi_\theta(B)$.
        \STATE Compute $\mathcal{L}_{\text{total}} = \mathcal{L}_{\text{pot}} + \lambda_{\text{soft}}\mathcal{L}_{\text{soft}} + \mathcal{L}_{\text{slack}}$.
        \STATE Update $\theta$ via $\nabla_\theta \mathcal{L}_{\text{total}}$.
    \ENDFOR
\ENDFOR

\FOR{epoch $e = 1$ to $E_{S2}$}
    \FOR{batch $B \in \mathcal{D}$}
        \STATE $\hat{\mathbf{y}} \leftarrow \pi_\theta(B)$
        \STATE $\mathbf{y}_0 \leftarrow \hat{\mathbf{y}}$ \COMMENT{Init projection}
        \WHILE{$\|\mathbf{h}(\mathbf{y}_k)\|_\infty > \epsilon$ \AND $k < I_\text{max}$}
            \STATE $\mathbf{y}_{k+1} \leftarrow \mathbf{y}_k - \mathbf{J}_{\mathbf{h}}^\dagger (\mathbf{y}_k) \mathbf{h}(\mathbf{y}_k)$
            \STATE $k \leftarrow k + 1$.
        \ENDWHILE
        \STATE Update $\theta$ via $\nabla_\theta (\mathcal{L}_{\text{pot}} + \lambda_{\text{proj}}\|\hat{\mathbf{y}}_0 - \mathbf{y}_k\|_2^2)$.
    \ENDFOR
\ENDFOR
\RETURN $\pi_\theta$
\end{algorithmic}
\end{algorithm}

\section{EXPERIMENTS}

\subsection{Experimental Setup}

\subsubsection{Environment and Data}
We designed challenging experimental scenarios to validate navigation in obstacle-dense unstructured terrains. The ego vehicle is modeled as a kinematic bicycle with a wheelbase $L=2.8$~m, width $w=1.9$~m, and maximum steering angle $\delta_{\text{max}}=0.70$ rad~\cite{polack2017kinematic}.
The environment spans a $40\times20$~m area, where $N_{\text{obs}}=8$ random quadrilateral obstacles are generated with side lengths $l \sim \mathcal{U}(1, 4)$~m and goals are randomly sampled within the target area on the right. Details are provided in Supplementary Material. A total of 200,000 scenarios were generated and partitioned into training, validation, and testing sets (6:3:1).

\subsubsection{Baselines and Metrics}
We compare our framework against four representative baselines. The implementation details and hyperparameters of each baseline are provided in Supplementary Material.
\begin{enumerate}
    \item Search-based: Hybrid A*~\cite{HybridAStar} utilizing Reeds-Shepp curves for kinematic feasibility.
    \item Sampling-based: Informed RRT*~\cite{gammell2014informed} with obstacle inflation (1.2~m).
    \item Optimization-based: NMPC~\cite{rawlings2020model} implemented via the CasADi framework\cite{andersson2019casadi} and the IPOPT solver\cite{wachter2006implementation}, sharing the exact kinematic model and differentiable obstacle representation as our method.
    \item Learning-based: IL trained on Hybrid A* expert demonstrations, and a Soft-Constraint network trained without the projection layer. This baseline shares the same network structure and parameters as our method but excludes the slack variables from the output layer.
\end{enumerate}

\begin{table*}[t]
  \centering
  \caption{Comparative Analysis of Path Planning Performance}\label{tab:comparison_static}
  \begin{threeparttable}
    \begin{tabular*}{0.9\textwidth}{@{\extracolsep{\fill}}llccccc}
      \toprule
      Category & Method & CT (s)$^{\dagger}$ & APL (m) & SR (\%) & $\mathcal{S}_{\text{kin}}$ & AFD (m) \\
      \midrule
      \multirow{2}{*}{Rule-based} 
        & Hybrid A* & 12.3392 & 35.57 & \textbf{98.80} & \textbf{0.9999} & \textbf{0.00} \\
        & Informed RRT* & 9.4949 & 34.59 & 86.26 & 0.9650 & \textbf{0.00} \\
      \midrule
      Optimization & NMPC & 1.5308 & 34.47 & 82.35 & 0.9969 & \textbf{0.00} \\
      \midrule
      \multirow{3}{*}{Learning-based} 
        & IL (Pure) & \textbf{0.0014} & 33.52 & 12.40 & 0.9921 & 0.57 \\
        & IL + Soft & \textbf{0.0014} & 33.45 & 54.30 & 0.9998 & 1.25 \\
        & Soft & \textbf{0.0014} & \textbf{33.31} & 55.22 & 0.9976 & 0.88 \\
        & Hard (Ours) & 0.0939 & \textbf{33.30} & 88.75 & 0.9885 & 0.53 \\
      \bottomrule
    \end{tabular*}
    \begin{tablenotes}
      \footnotesize
      \item[$\dagger$] CT for Learning-based methods is measured on Jetson Orin NX, while other baselines are measured on the Intel i9-14900KF CPU.
      \item \textbf{Bold} values indicate the best results for each metric. 
    \end{tablenotes}
  \end{threeparttable}
\end{table*}
\begin{figure*}[t]
    \centering
    \setlength{\tabcolsep}{1pt} 
    \begin{minipage}{\textwidth}
        \raggedleft
        \includegraphics[width=0.5\linewidth]{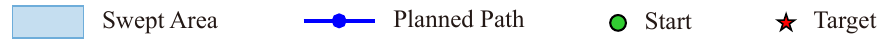}
    \end{minipage}
    \begin{tabular}{c c c c c c}
      \subfloat{\includegraphics[width=0.162\linewidth]{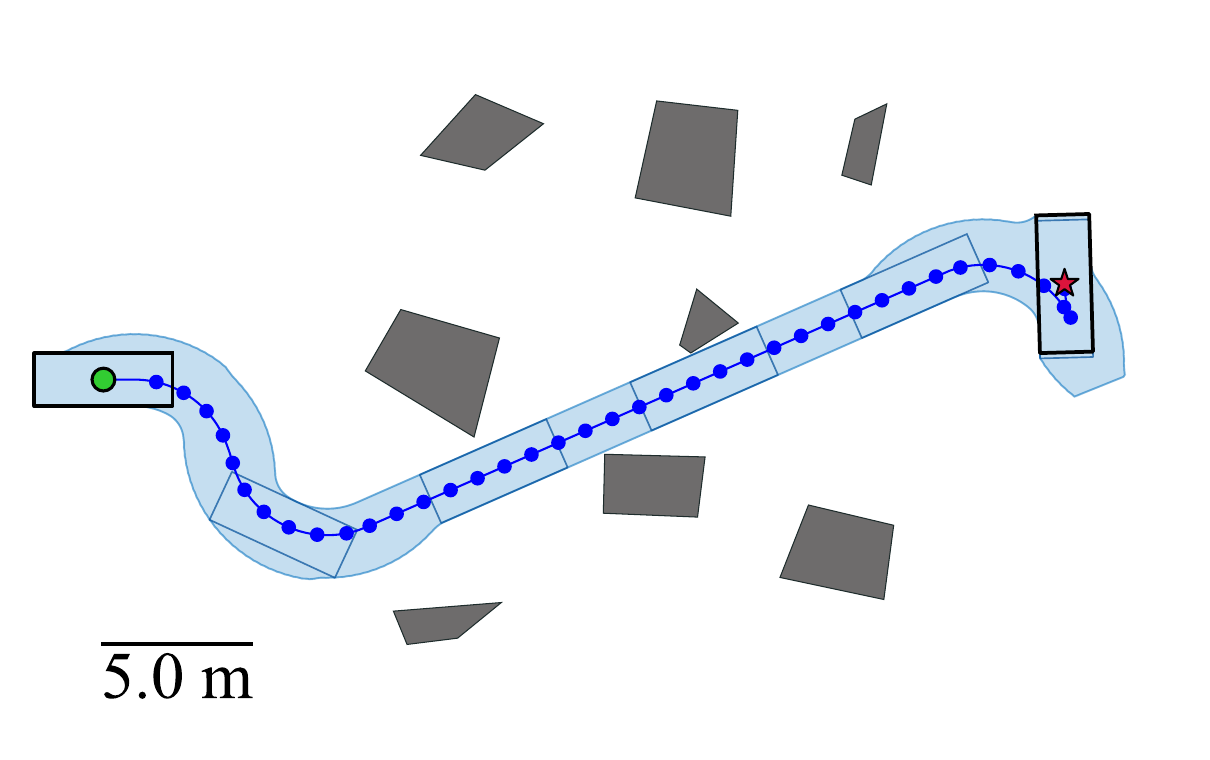}} &
      \subfloat{\includegraphics[width=0.162\linewidth]{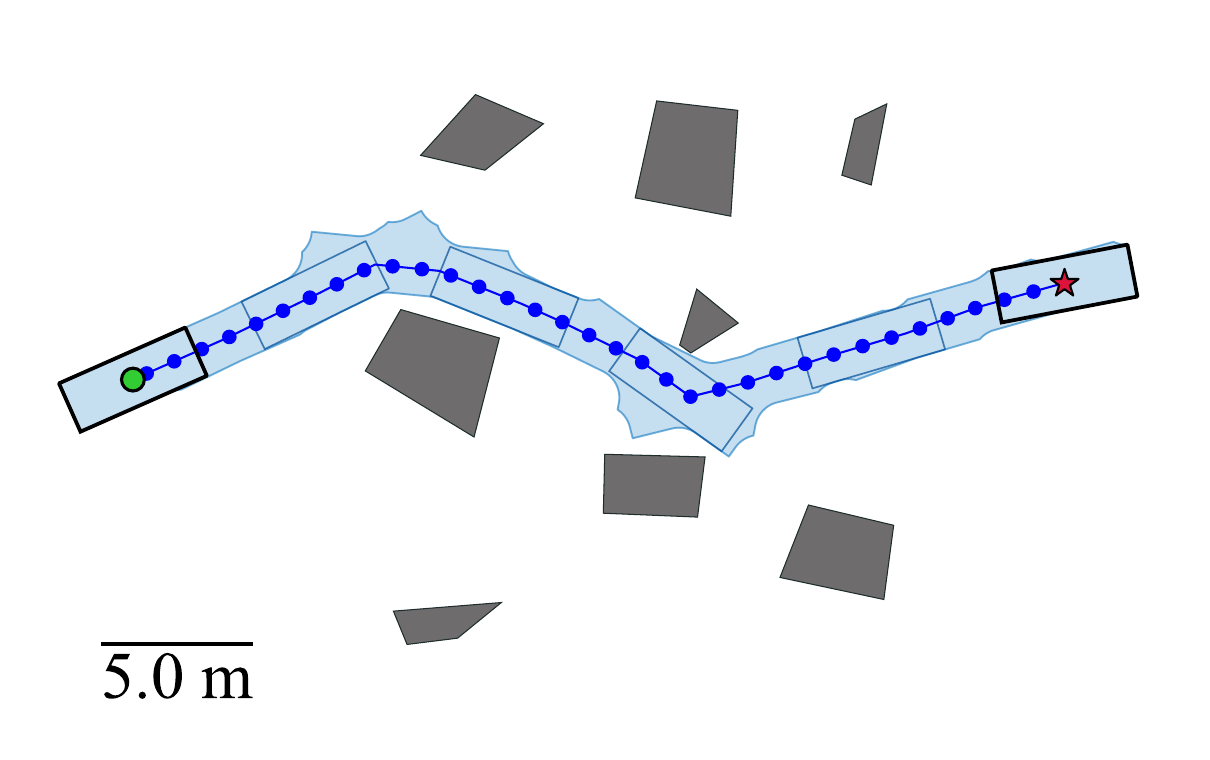}} &
      \subfloat{\includegraphics[width=0.162\linewidth]{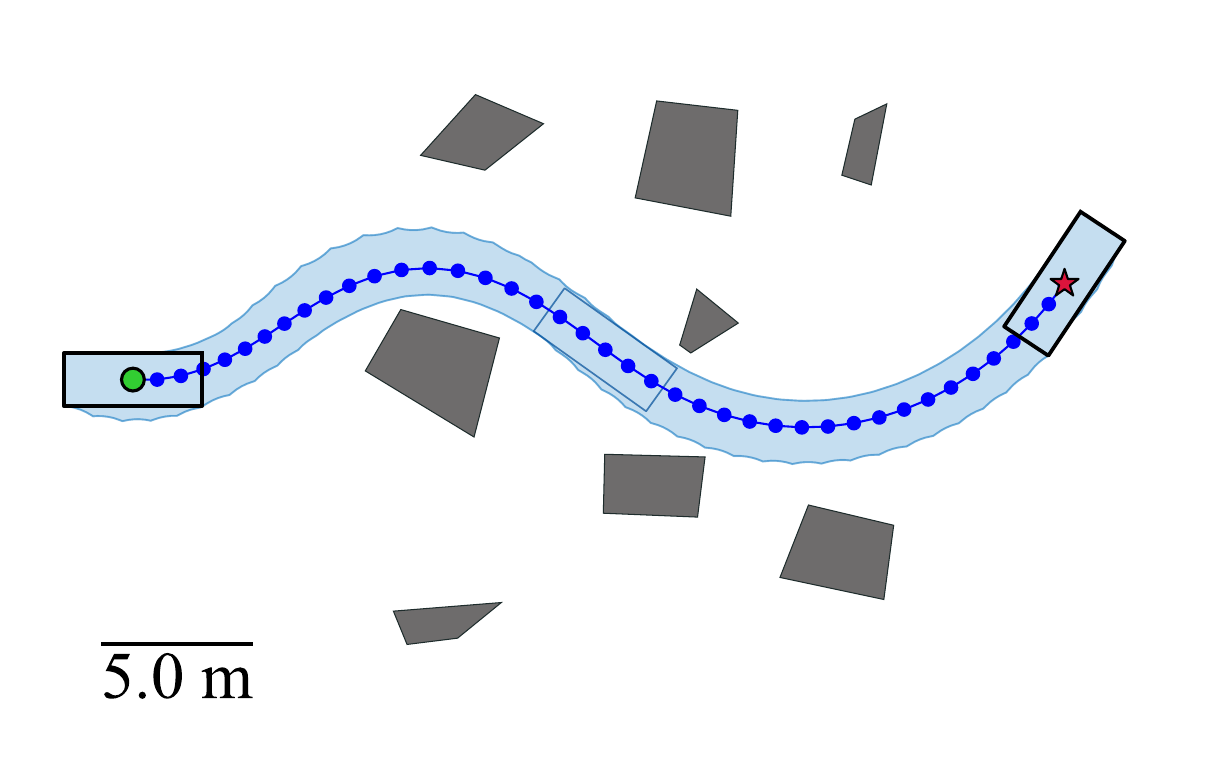}} &
      \subfloat{\includegraphics[width=0.162\linewidth]{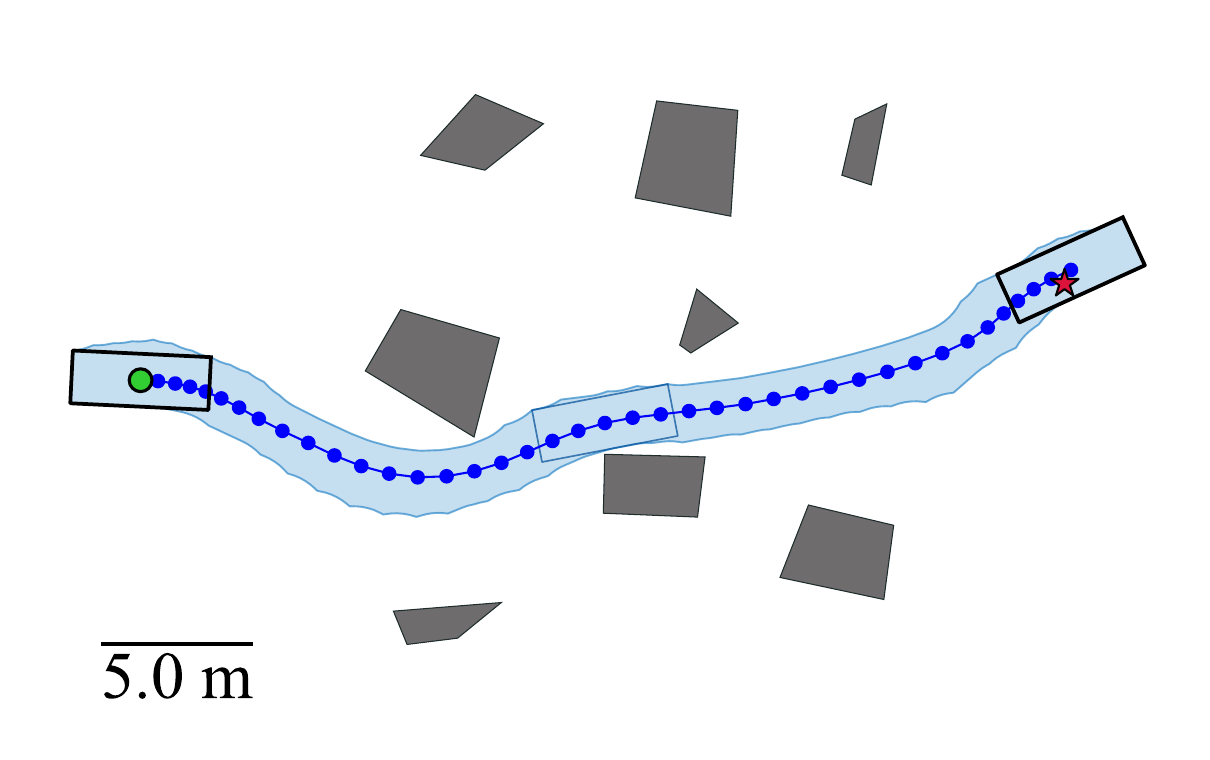}} &
      \subfloat{\includegraphics[width=0.162\linewidth]{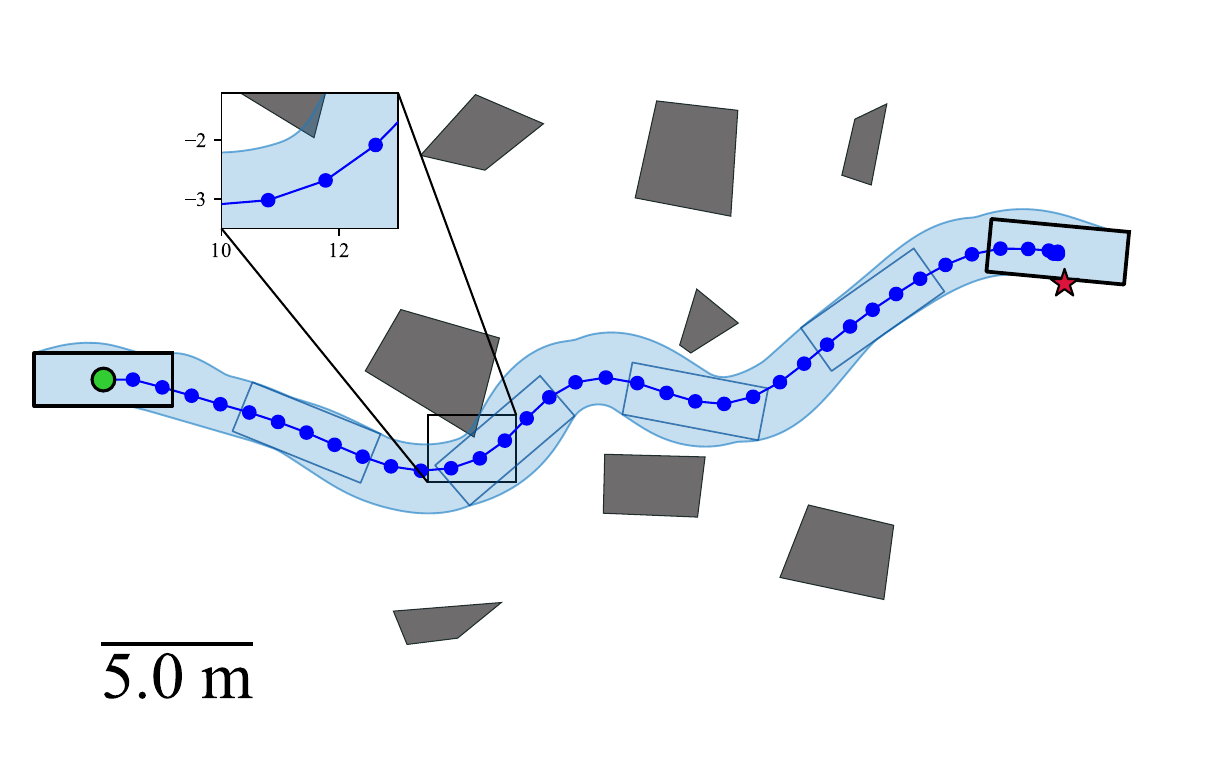}} &
      \subfloat{\includegraphics[width=0.162\linewidth]{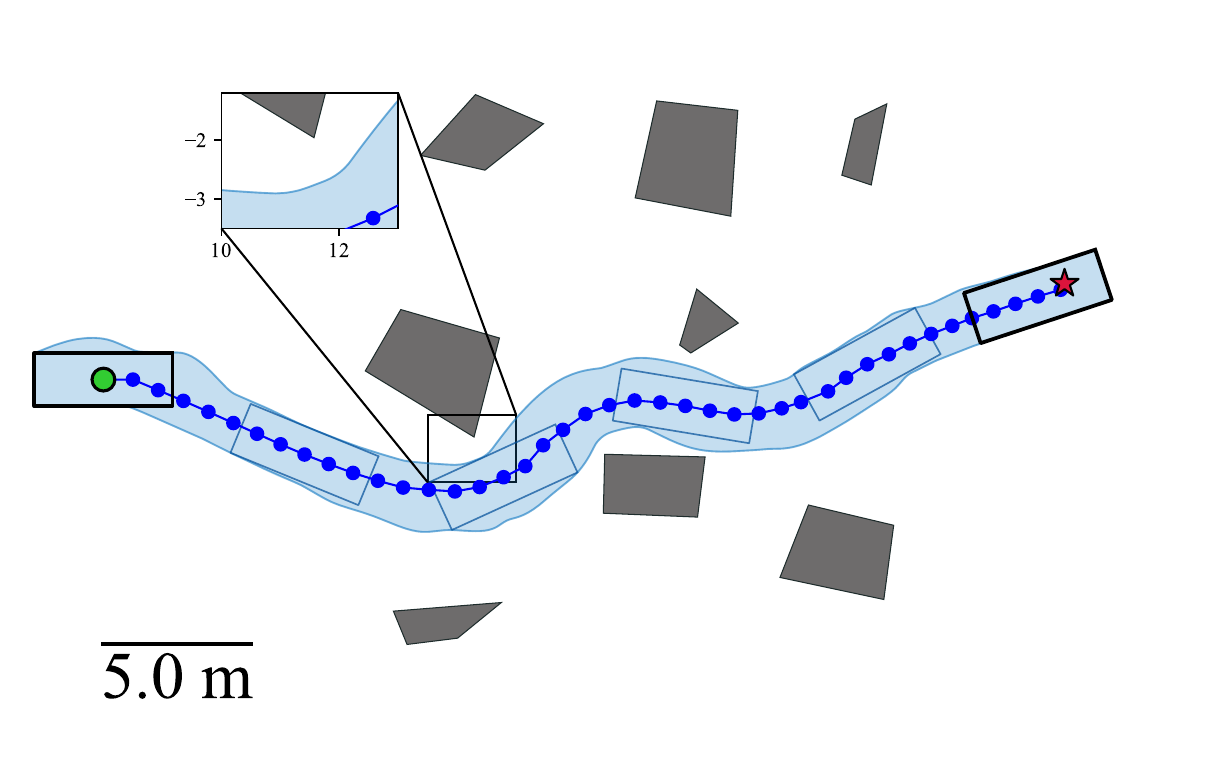}} \\
      \subfloat{\includegraphics[width=0.162\linewidth]{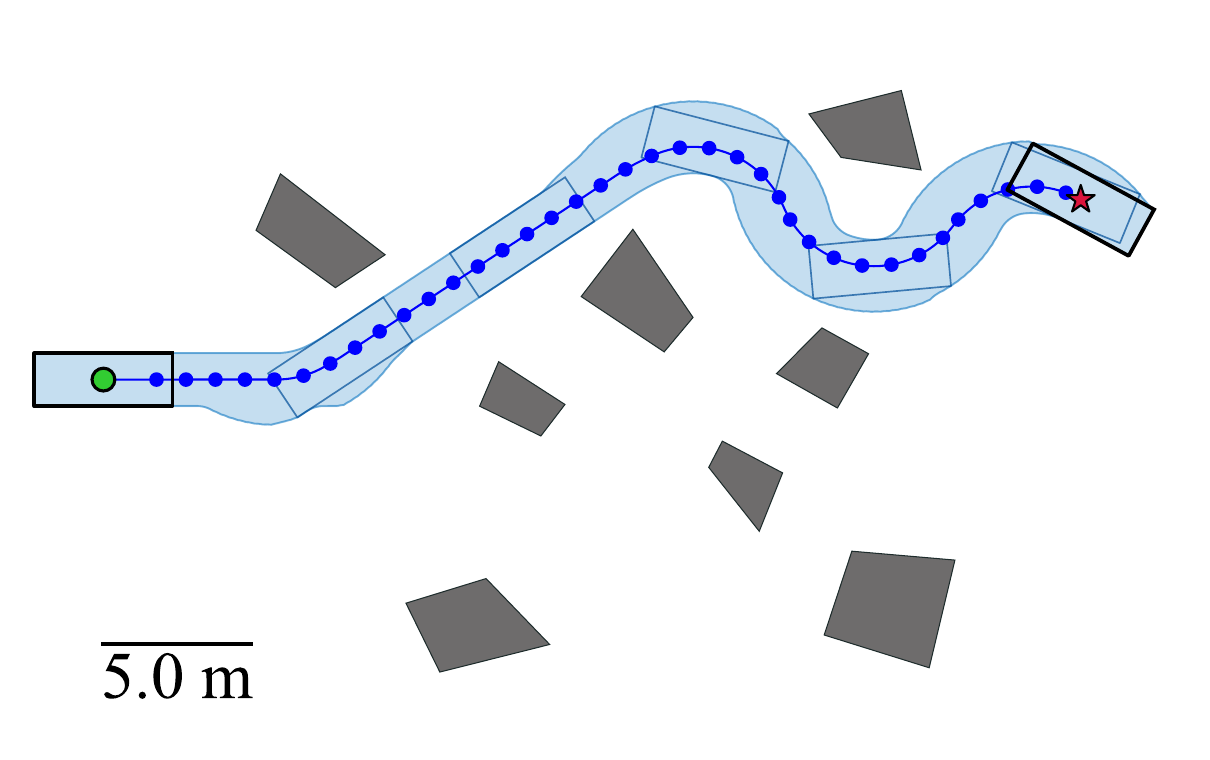}} &
      \subfloat{\includegraphics[width=0.162\linewidth]{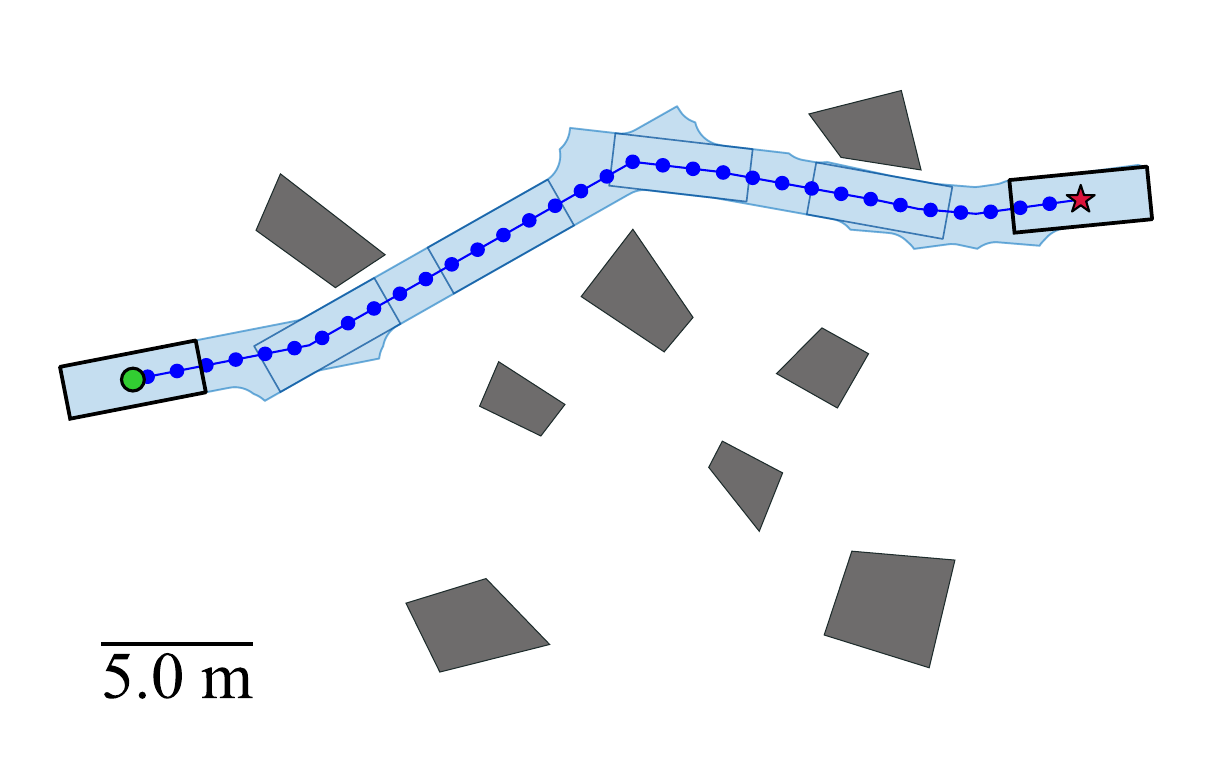}} &
      \subfloat{\includegraphics[width=0.162\linewidth]{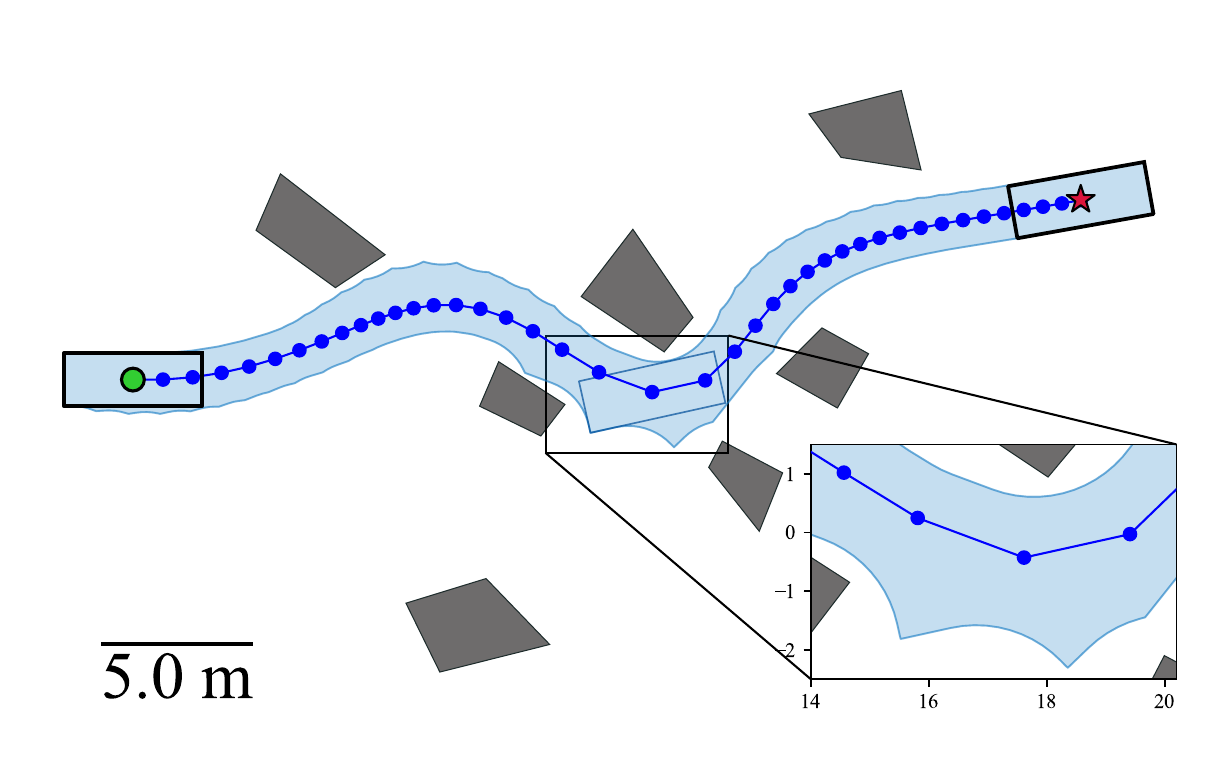}} &
      \subfloat{\includegraphics[width=0.162\linewidth]{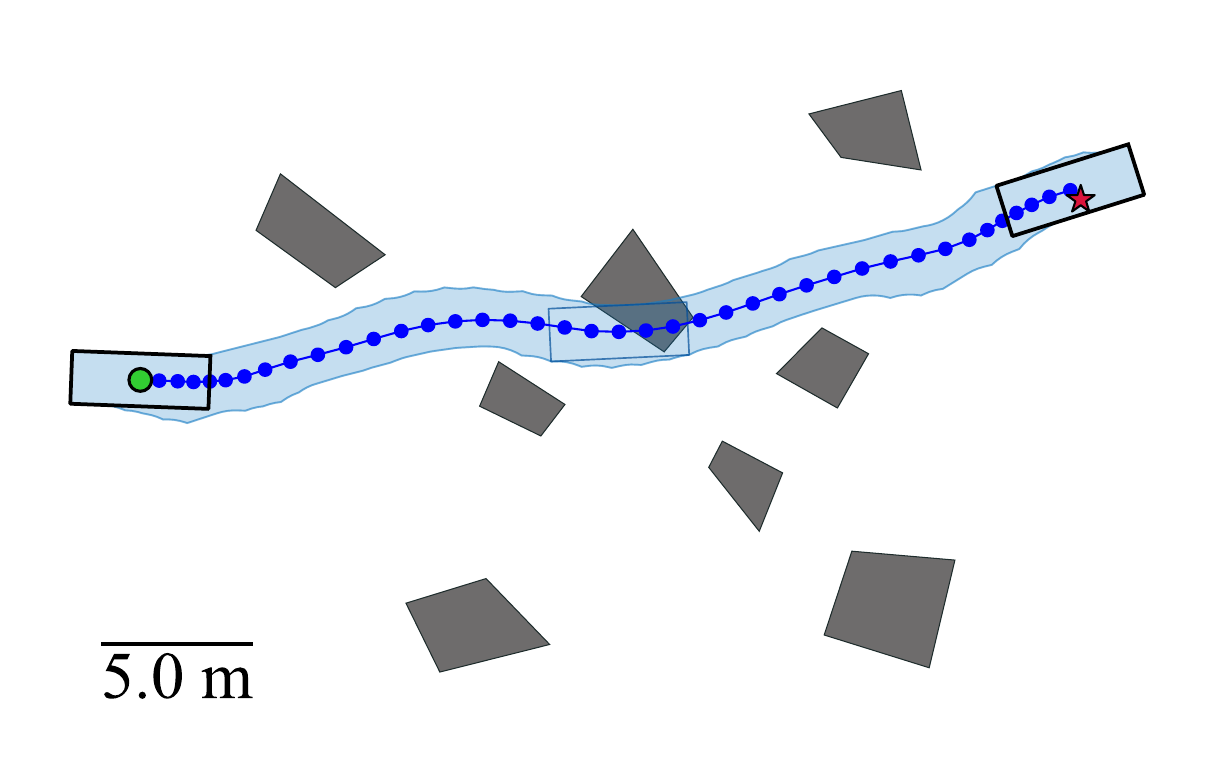}} &
      \subfloat{\includegraphics[width=0.162\linewidth]{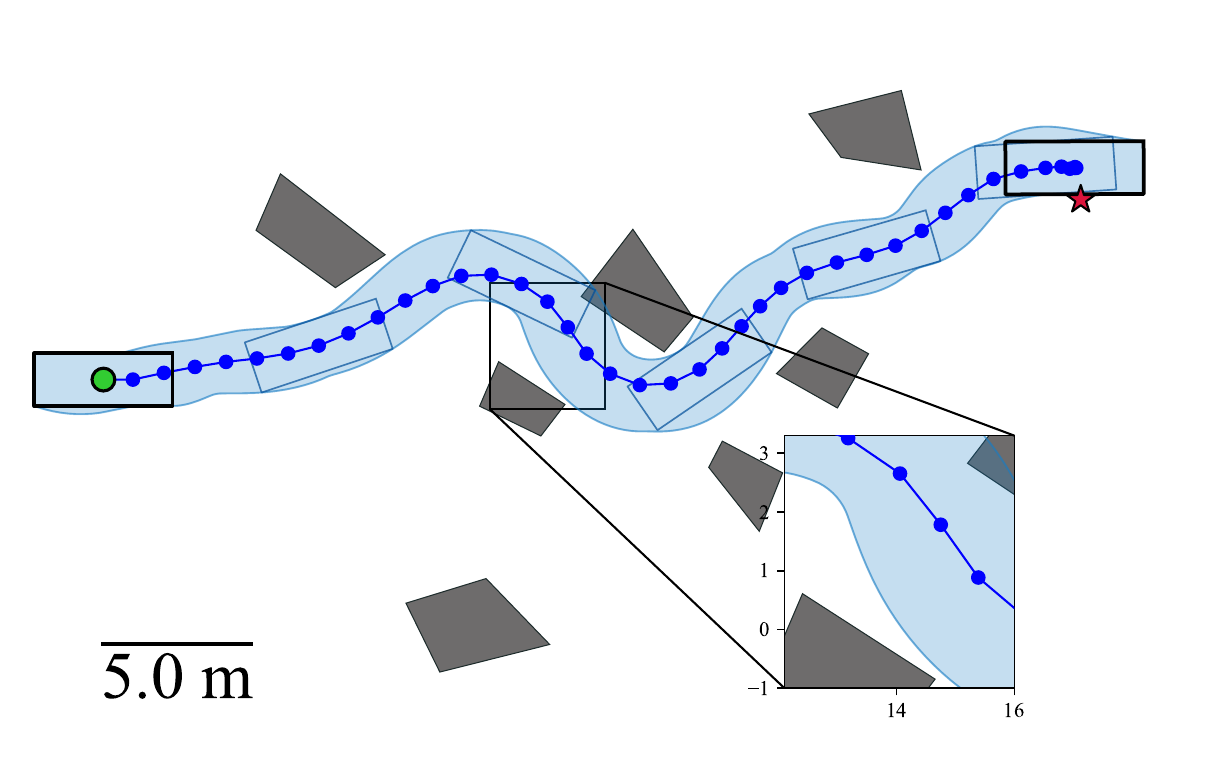}} &
      \subfloat{\includegraphics[width=0.162\linewidth]{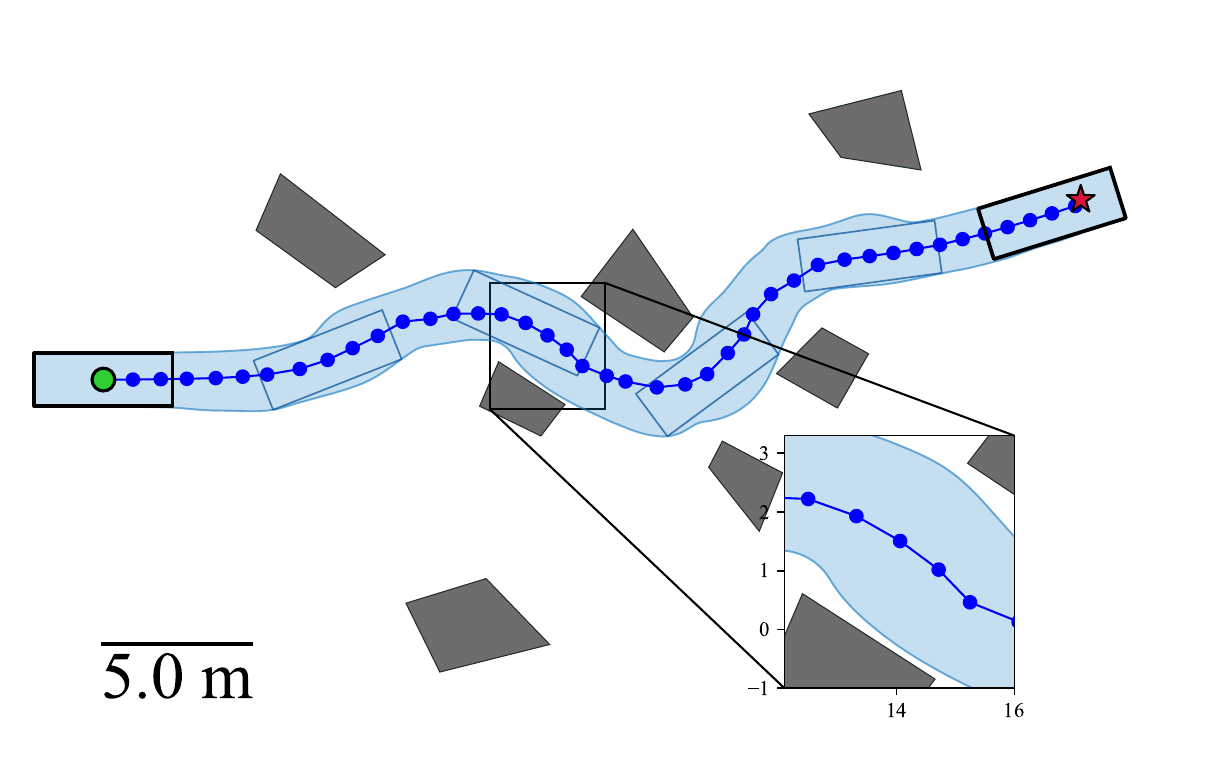}} \\
      \small \textbf{Hybrid A*} & 
      \small \textbf{Informed RRT*} & 
      \small \textbf{NMPC} & 
      \small \textbf{IL+Soft} & 
      \small \textbf{Soft} & 
      \small \textbf{Hard (Ours)} \\

    \end{tabular}

    \caption{Qualitative comparison of generated paths in two narrow-corridor scenarios (top and bottom rows). As highlighted in the zoomed-in regions: the NMPC baseline exhibits sparse waypoint distribution, which poses a risk of collision during actual execution due to interpolation errors. The comparison between the Soft and Hard methods demonstrates the effectiveness of the proposed hard constraint formulation in strictly guaranteeing collision-free swept areas within tight spaces.}\label{fig:visual_comparison}
\end{figure*}
Performance is evaluated using static metrics: Success Rate (SR; collision-free arrival within 1.0~m), Average Path Length (APL), Computation Time (CT), and Kinematic Compliance ($\mathcal{S}_{\text{kin}} = \sum_{t=1}^{T} \min ( \kappa_{\text{max}}/\kappa_t, 1 )/T$).

To validate physical feasibility, we conduct closed-loop tracking experiments in CARLA~\cite{dosovitskiy2017carla}. The planned paths are tracked by a PID controller. Detailed configurations are provided in Supplementary Material. We report Tracking Success Rate (TSR), Cross-Track Error (RMSE CTE), and Steering Smoothness ($\mathcal{J}_{\text{smooth}} = \sum_{t=0}^{T-1} | \delta_{t+1} - \delta_t |/T$).

\subsubsection{Implementation Details}
The policy network is a five-layer MLP with hidden dimensions $[128, 256, 512, 768, 512]$ and ReLU activations trained using Adam\cite{kingma2015adam}. Training consists of a soft-constraint pre-training stage (400 epochs, $lr=10^{-4}$, batch size 256) followed by a hard-constraint refinement stage (40 epochs, $lr=10^{-5}$, batch size 64). During inference, the hard projection layer uses a tolerance of $\epsilon = 10^{-3}$ with a maximum 50 iterations. Training utilized an Intel i9-14900KF CPU and NVIDIA RTX 2080 Ti GPU. For inference, learning-based methods were deployed on an NVIDIA Jetson Orin NX, while traditional planners were evaluated on the Intel i9-14900KF CPU.

\subsection{Comparative Analysis of Planning Performance}

Table \ref{tab:comparison_static} presents a quantitative evaluation across 20,000 scenarios. The results reveal that existing methods face distinct limitations. Search-based Hybrid A* excels in completeness, achieving a 98.80\% success rate. However, this comes at the cost of prohibitive latency ($>12$ s), caused by the exponential expansion of motion primitives in the search space. Conversely, pure learning methods provide negligible computation times ($\approx 1.4$ ms) but struggle with safety. Lacking explicit geometric constraints, their success rate drops below 55\%. As for optimization-based NMPC, while it incorporates constraints, it achieves only an 82.35\% success rate. This performance cap arises because the solver frequently gets trapped in local minima due to its high sensitivity to initialization.

Our framework strikes a superior balance between computational efficiency and safety guarantees. By integrating a differentiable projection layer designed to explicitly enforce hard constraints, our method achieves an 88.75\% success rate, significantly outperforming NMPC. Unlike Informed RRT*, which generates stochastic and often jagged paths, our method produces deterministic and smoother paths, as evidenced by the superior Average Path Length (33.30~m).
Computationally, our method maintains an efficient inference latency of 0.0939 s on the Jetson Orin NX, achieving an acceleration of over two orders of magnitude compared to Hybrid A* running on Intel i9-14900KF. This efficiency explicitly validates the method's suitability for resource-constrained embedded systems.

\subsection{Dynamic Validation and Path Tracking}
Table \ref{tab:comparison_dynamic} summarizes the dynamic validation results obtained from 2,000 testing scenarios.

\begin{table*}[t]
  \centering
  \caption{Comparative Analysis of Path Tracking Performance}\label{tab:comparison_dynamic}
  \begin{threeparttable}
    \begin{tabular*}{0.9\linewidth}{@{\extracolsep{\fill}}lccccccc}
      \toprule
      Method & RMSE CTE (m) & Max CTE (m) & AHE (°) & DFR (\%) & $\mathcal{J}_{\text{smooth}}$ & TSR (\%) & SR Drop (\%)\\
      \midrule
        Hybrid A* & 0.2505 & 0.6102 & 5.122 & 96.25 & 0.0335 & \textbf{94.65} & \textbf{4.15}\\
        Informed RRT* & 0.2715  & 0.6090  & 6.619  & 94.70  & \textbf{0.0311} & 62.55  & 23.71\\ 
        NMPC & \textbf{0.1041}  & \textbf{0.2441}  & 4.367  & 98.53  & 0.0312 & 74.48  & 7.87\\ 
        IL+Soft & 0.1473  & 0.3214  & \textbf{4.184}  & \textbf{99.56}  & 0.0429  & 44.35 & 9.95\\ 
        Soft & 0.1860  & 0.4149  & 6.276  & 98.02  & 0.0371 & 41.70 & 12.45\\ 
        Hard (Ours) & 0.1982  & 0.4544  & 5.823  & 97.70  & 0.0377 & 80.40 & 8.35\\ 
      \bottomrule
    \end{tabular*}
    \begin{tablenotes}
      \footnotesize
      \item \textbf{Bold} values indicate the best results for each metric. 
    \end{tablenotes}
  \end{threeparttable}
\end{table*}

\begin{figure*}[t]
    \centering
    \subfloat[Hybrid $A^{*}$]{
        \includegraphics[width=0.3\linewidth]{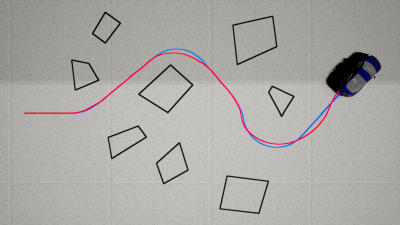}
        \label{fig:track_hybrid}
    }
    \subfloat[Informed RRT*]{
        \includegraphics[width=0.3\linewidth]{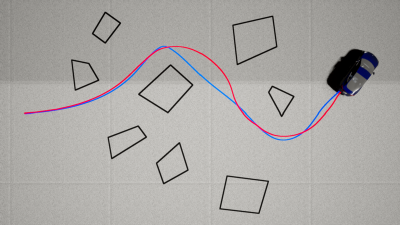}
        \label{fig:track_rrt}
    }
    \subfloat[NMPC]{
        \includegraphics[width=0.3\linewidth]{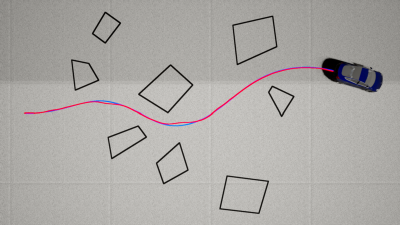}
        \label{fig:track_nmpc}
    }
    \\ % 强制换行，保持3x2布局
    \subfloat[IL + Soft]{
        \includegraphics[width=0.3\linewidth]{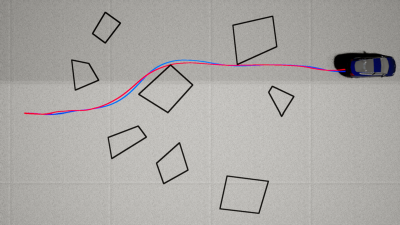}
        \label{fig:track_il}
    }
    \subfloat[Soft]{
        \includegraphics[width=0.3\linewidth]{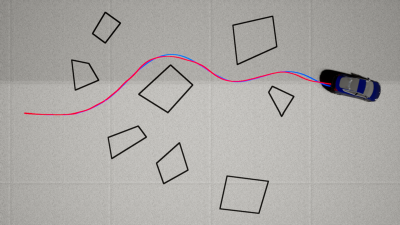}
        \label{fig:track_soft}
    }
    \subfloat[Ours (Hard)]{
        \includegraphics[width=0.3\linewidth]{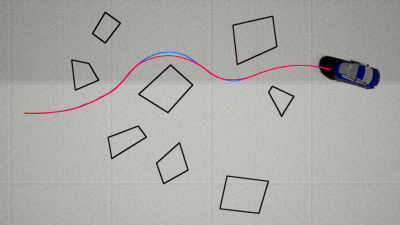}
        \label{fig:track_ours}
    }
    \caption{Visual comparison of path tracking in CARLA. Obstacles are depicted as black polygons, while blue and red lines denote planned paths and actual tracking trajectories, respectively. Note that Hybrid $A^*$ (a) exhibits control oscillations during continuous turns, and Informed RRT* (b) suffers from large tracking errors due to infeasible sharp turns. The Soft baselines (d, e) fail to avoid obstacles effectively. In contrast, both NMPC (c) and our method (f) demonstrate high kinematic feasibility while strictly ensuring operational safety.}
    \label{fig:four_algo_tracking}
\end{figure*}
\subsubsection{Tracking Robustness and Feasibility Gap}
As shown in Table~\ref{tab:comparison_dynamic}, Hybrid A* establishes the benchmark with a Tracking Success Rate (TSR) of 94.65\%. In contrast, pure learning baselines (IL+Soft and Soft) fail to ensure safety ($<45\%$ TSR), and NMPC (74.48\%) frequently stagnates in local minima.
Our framework achieves a TSR of 80.40\%. A key metric here is the \textit{SR Drop}, which quantifies the degradation from static planning to dynamic execution. Informed RRT* suffers a drastic drop of 23.71\%, indicating that its geometrically feasible paths are often dynamically infeasible. Conversely, our method maintains a minimal SR Drop of 8.35\%. By integrating kinematic constraints into the projection process, our method guarantees dynamically feasible paths that result in superior tracking performance.

\begin{figure}[htbp]
    \centering
    \includegraphics[width=0.5\textwidth]{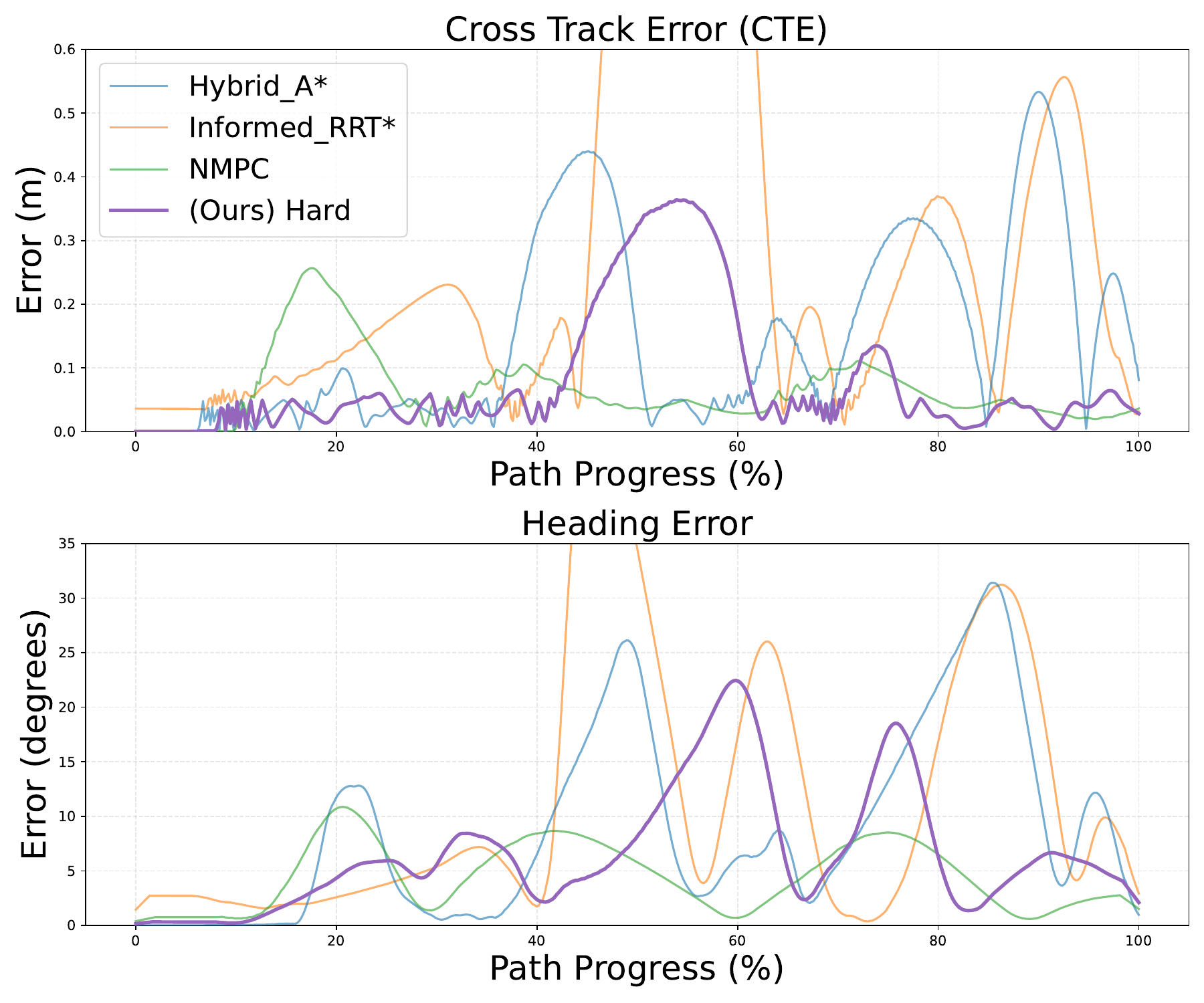}
    \caption{Tracking error analysis. Our method (black line) maintains low cross-track and heading errors throughout the trajectory, mitigating the oscillations seen in Hybrid A* and the spikes in RRT*.}\label{fig:error}
\end{figure}
\subsubsection{Control Accuracy and Stability}
In terms of tracking precision, NMPC achieves the lowest RMSE CTE (0.1041~m) due to its explicit model-based formulation. While Soft baselines yield low tracking errors, this is primarily because they generate overly smooth paths that violate obstacle constraints, resulting in high failure rates despite precise tracking.

Our method maintains a competitive RMSE CTE of 0.1982~m. Notably, this accuracy surpasses both Hybrid A* (0.2505~m) and Informed RRT* (0.2715~m). As visualized in Fig.~\ref{fig:four_algo_tracking}(a), Hybrid A* relies on Reeds-Shepp primitives, which introduce curvature discontinuities at junction points. Our method exhibits a high Dynamic Feasibility Ratio (DFR: 97.7\%) and superior smoothness. This confirms the high physical feasibility of the paths generated by our method.

\subsection{Ablation Studies and Sensitivity Analysis}
To comprehensively evaluate the robustness of the proposed framework and verify the contribution of its core modules, we investigated the impact of three critical factors: $\lambda_{\text{soft}}$ in the pre-training stage, $I_\text{max}$ of the hard constraint projection layer, and the fundamental necessity of the G-APF.

% --- Table 1: Soft Constraint ---
\begin{table}[htbp]
  \centering
  \caption{Ablation Study of Soft Constraint Weighting ($\lambda_{\text{soft}}$)}
  \label{tab:ablation_soft}
  % \resizebox{\columnwidth}{!}{ % 自动缩放表格以适应单栏宽度
    \begin{tabular}{l c c c c c}
      \toprule
      $\lambda_{\text{soft}}$ & CT (s) & APL (m) & SR (\%) & $\mathcal{S}_{\text{kin}}$ & AFD (m) \\
      \midrule
        0.5 & 0.0014 & 33.18 & 48.82 & 0.9975 & 1.0250 \\
        0.8 & 0.0014 & 33.34 & 50.04 & 0.9971 & 1.0560 \\
        1.0 & 0.0014 & 33.36 & 52.22 & 0.9974 & 1.0224 \\
        1.2 & 0.0014 & 33.31 & 55.22 & 0.9976 & 0.8785 \\
        1.5 & 0.0014 & 13.61 & 0.00 & 0.7825 & 18.9200 \\
      \bottomrule
    \end{tabular}
  % }
\end{table}

\subsubsection{Impact of $\lambda_{\text{soft}}$}
The weight $\lambda_{\text{soft}}$ balances obstacle avoidance and goal attraction during pre-training. As shown in Table~\ref{tab:ablation_soft}, increasing $\lambda_{\text{soft}}$ from 0.5 to 1.2 steadily improves SR, indicating that stronger penalties effectively encourage obstacle-aware representations. However, a threshold exists; at $\lambda_{\text{soft}} = 1.5$, the planner becomes overly conservative, trapping the agent in local minima (0\% SR) due to excessive repulsive forces. Consequently, we selected $\lambda_{\text{soft}} = 1.2$ as the optimal balance between safety awareness and goal reachability.

% --- Table 2: Iterations ---
\begin{table}[htbp]
  \centering
  \caption{Ablation Study of Projection Iterations ($I_\text{max}$)}
  \label{tab:ablation_iter}
    \begin{tabular}{l c c c c c}
      \toprule
      $I_\text{max}$ & CT (s) & APL (m) & SR (\%) & $\mathcal{S}_{\text{kin}}$ & AFD (m) \\
      \midrule
        20 & 0.0427 & 33.36 & 87.01 & 0.9872 & 0.5081 \\
        50 & 0.0939 & 33.32 & 88.75 & 0.9885 & 0.5345 \\
        80 & 0.1245 & 33.30 & 89.58 & 0.9889 & 0.5465 \\
        100 & 0.1363 & 33.30 & 89.84 & 0.9890 & 0.5719 \\
      \bottomrule
    \end{tabular}
\end{table}

\subsubsection{Impact of $I_\text{max}$}
Table~\ref{tab:ablation_iter} examines the trade-off between solvability and computation time. Increasing $I_\text{max}$ from 20 to 50 significantly boosts SR (87.01\% to 88.75\%) and reduces AFD. However, further increasing iterations to 100 yields diminishing returns ($\approx 1\%$ SR gain) at the cost of increased computational overhead. To satisfy the real-time constraint of autonomous navigation in unstructured environments, we fixed $I_\text{max} = 50$ as the optimal compromise between kinematic feasibility and efficiency.

% --- Table 3: G-APF ---
\begin{table}[ht]
  \centering
  \caption{Ablation Study on Global-Guided Potential Field}
  \label{tab:ablation_gapf}
    \begin{tabular}{l c c c c c}
      \toprule
      Method & CT (s) & APL (m) & SR & $\mathcal{S}_{\text{kin}}$ & AFD (m) \\
      \midrule
      w/o G-APF & 0.0010 & 12.98 & 0.00\% & 0.9918 & 21.95 \\
      Hard (Ours)      & 0.2079 & 33.30 & 88.75\% & 0.9885 & 0.53 \\
      \bottomrule
    \end{tabular}
\end{table}

\subsubsection{Impact of G-APF} To verify the necessity of the proposed global guidance, we replaced the G-APF with a standard Euclidean distance heuristic. Table~\ref{tab:ablation_gapf} reveals that the Euclidean baseline yields a 0.00\% SR. This zero success rate is attributed to the agent getting trapped in local minima and consequently failing to reach the goal, rather than collisions, as evidenced by the large AFD of 21.95~m. Note that the second-stage projection was omitted for this baseline because the hard constraint layer cannot converge without a coarse-feasible initial guess.

\section{CONCLUSION}
This paper proposes a self-supervised path planning framework designed for embedded mechatronics in unstructured environments. By coupling global topological guidance with differentiable constraint projection, the proposed architecture achieves data-efficient learning and explicit runtime safety without requiring expert demonstrations.
Extensive experiments and embedded deployment on an NVIDIA Jetson Orin NX validate the system's performance, achieving an 88.75\% success rate with a real-time latency of 94 ms.

Future work will extend this embedded intelligence framework to dynamic environments by integrating motion prediction into Spatio-Temporal Potential Fields, enabling safe interaction with moving agents. Additionally, we plan to investigate hybrid search-learning strategies to provide robust warm starts for the projection solver, further enhancing the system's resilience and convergence reliability in extreme mechatronic operating conditions.

\section*{Acknowledgments}
This work was supported in part by National Natural Science Foundation of China under Grant 52275564, U24A20109, and by Beijing Natural Science Foundation under Grant  QY24251. 

The authors would like to thank Chengyun Ju, Donggang Sang, and Bingtao Zhang from Hirain Inc. for their valuable discussions and technical support during the course of this work.

\bibliographystyle{unsrt}
\bibliography{references}

\end{document}